\documentclass{article}

\usepackage{arxiv}

\usepackage[utf8]{inputenc} % allow utf-8 input
\usepackage[T1]{fontenc}    % use 8-bit T1 fonts
\usepackage{hyperref}       % hyperlinks
\usepackage{url}            % simple URL typesetting
\usepackage{booktabs}       % professional-quality tables
\usepackage{amsfonts}       % blackboard math symbols
\usepackage{nicefrac}       % compact symbols for 1/2, etc.
\usepackage{microtype}      % microtypography
\usepackage{lipsum}
\usepackage{graphicx}
\usepackage{amsmath}
\usepackage{indentfirst}
\usepackage{caption}
\usepackage{subcaption}
\usepackage[inkscapelatex=false]{svg}
\usepackage[left]{lineno}       % ← the key package
\modulolinenumbers[1]           % number every fifth line (set to 1 for every line)

\graphicspath{ {./images/} }
\title{Unsupervised Document and Template Clustering using Multimodal Embeddings}

\author{
 Phillipe R. Sampaio\thanks{Corresponding author:
 \texttt{phillipe.rodriguessampaio@bnpparibas.com}} \\
  BNP Paribas Cardif\\
  Nanterre, France \\
   \And
 Helene Maxcici \\
  BNP Paribas\\
  Paris, France \\
}

\begin{document}
%\linenumbers 
\setlength\parindent{15pt}
\maketitle

\begin{abstract}
We study unsupervised clustering of documents at both the category and template levels using frozen multimodal encoders and classical clustering algorithms. We systematize a model-agnostic pipeline that (i) projects heterogeneous last-layer states from text--layout--vision encoders into token-type--aware document vectors and (ii) performs clustering with centroid- or density-based methods, including an HDBSCAN + $k$-NN assignment to eliminate unlabeled points. We evaluate eight encoders (text-only, layout-aware, vision-only, and vision--language) with $k$-Means, DBSCAN, HDBSCAN + $k$-NN, and BIRCH on five corpora spanning clean synthetic invoices, their heavily degraded print-and-scan counterparts, scanned receipts, and real identity and certificate documents. The study reveals modality-specific failure modes and a robustness--accuracy trade-off, with vision features nearly solving template discovery on clean pages while text dominates under covariate shift, and fused encoders offering the best balance. We detail a reproducible, oracle-free tuning protocol and the curated evaluation settings to guide future work on unsupervised document organization.
\end{abstract}


\section{Introduction}
\label{sec:introduction}

Document clustering groups documents into coherent sets without labels and underpins retrieval, management, topic discovery, and modern \emph{intelligent document processing} (IDP). Its importance grows with the scale of digital collections \cite{Cozzolino2022}. Early work used content-only features—bag-of-words with Frequency-Inverse Document Frequency (TF–IDF) weighted vectors—clustered by $k$-means or hierarchical methods \cite{Yan2017}. These approaches are effective for coarse topical structure but weak at capturing semantics.

Deep learning has advanced document representations. For example, autoencoders can learn compact embeddings \cite{Diallo2021}. Pre-trained language models such as BERT yield contextualized features, typically aggregated with the \texttt{[CLS]} token or with mean pooling to form document-level vectors \cite{Devlin2019,Liu2019}. Building on this trend, recent studies use large language model (LLM) embeddings directly for clustering and even employ generation to propose cluster centroids \cite{PETUKHOVA2025100,kLLMmeans2025}. However, many documents carry a salient layout and visual signals beyond text. Invoices, purchase orders, and reports differ not only by category but also by template and page structure. Exploiting these cues is critical in IDP and document layout analysis, whereas text-only pipelines may conflate structurally distinct documents with similar wording.

\begin{figure}[ht]
\centering
\captionsetup[subfigure]{justification=centering}
% ---------------------------------------------------- 1st row
\begin{subfigure}[t]{0.48\textwidth}
  \centering
  \includegraphics[width=\linewidth]{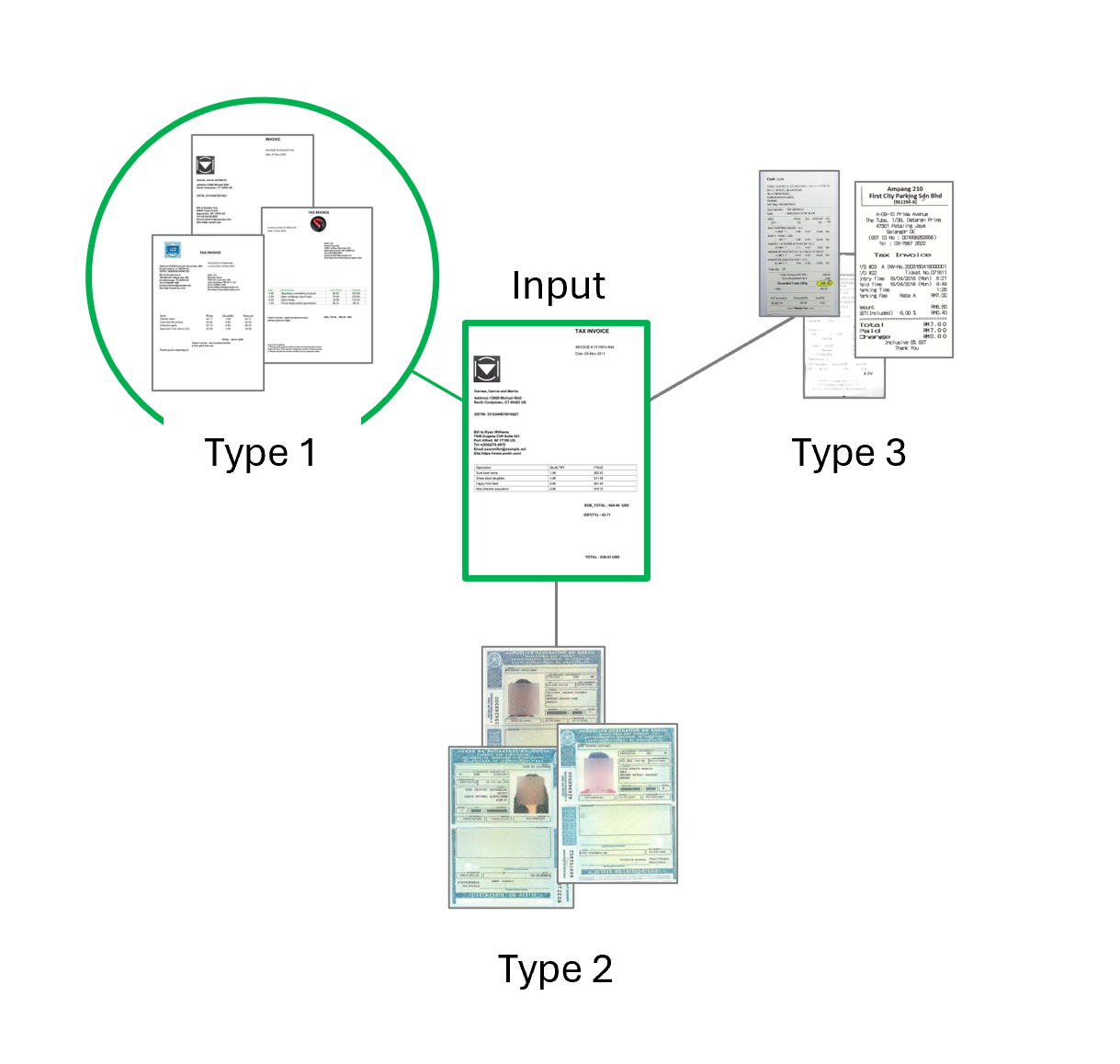}
  \caption{Clustering at the document level}
  \label{fig:doc_level_clustering}
\end{subfigure}
\hspace{0.02\textwidth} % Small horizontal space
\begin{subfigure}[t]{0.48\textwidth}
  \centering
  \includegraphics[width=\linewidth]{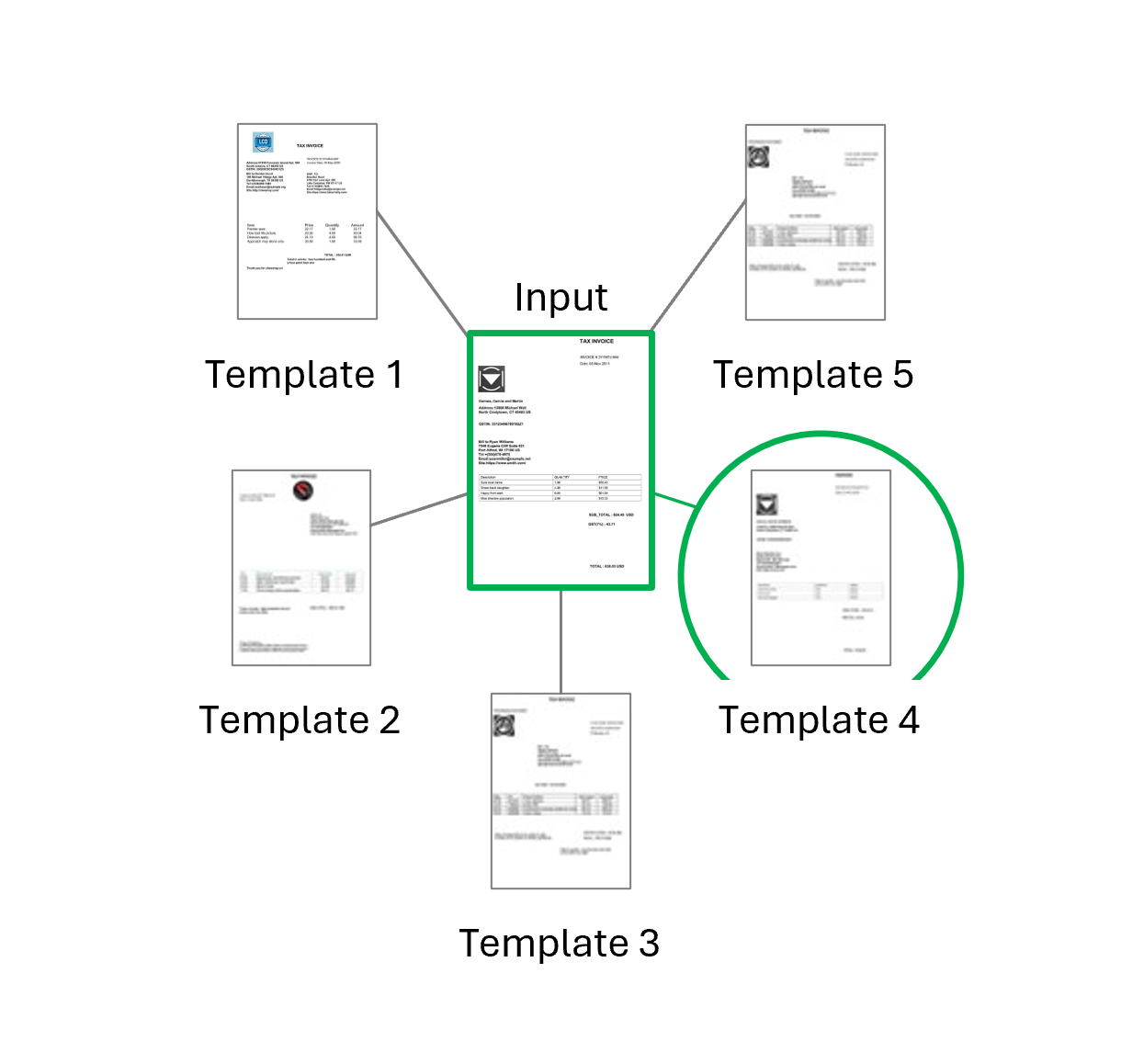}
  \caption{Clustering at the template level}
  \label{fig:template_level_clustering}
\end{subfigure}

\caption{Illustration of two levels of document clustering. In (a), documents are grouped based on their type, such as invoices, ID cards, and receipts. In (b), documents of the same type (e.g., invoices) are further clustered by their specific templates.}
\label{fig:types_of_clustering}
\end{figure}

In this paper, we address this limitation by proposing a document clustering approach that uses multimodal embeddings. These embeddings are specifically designed to capture not only textual content but also layout structures and visual characteristics inherent in documents. By integrating these rich representations into standard clustering algorithms such as $k$-Means \cite{forgy1965, lloyd1982, ahmed2020}, DBSCAN \cite{ester1996}, a combination of HDBSCAN \cite{campello2013} and $k$-NN \cite{FixHodges1951}, and BIRCH \cite{birch1996}, we aim to achieve a more comprehensive understanding of documents, facilitating clustering not only at the document-type level, but also at the template level within each type. For example, our approach can differentiate between distinct invoice templates from the same vendor or identify various versions of a financial report. Figure~\ref{fig:types_of_clustering} illustrates the two granularity levels considered in our document clustering approach. At the document level (Figure~\ref{fig:doc_level_clustering}), documents are grouped according to their overall type, such as invoices, ID cards, and receipts. At a finer granularity, the template-level clustering (Figure~\ref{fig:template_level_clustering}) separates documents of the same type based on variations in their layout or structure.

Template-level clustering presents significant advantages for IDP systems. In production environments, it allows for effective monitoring of machine learning models by enabling timely detection of new document templates, thus proactively identifying potential data drift scenarios. Additionally, in information extraction tasks, it helps verify whether the target information is indeed present in a document, as its availability is often contingent on the specific document template. This capability substantially improves the accuracy and reliability of information extraction processes. Template-level clustering also enables the automatic generation of structured metadata that reflects both the layout and content of documents. This significantly enhances indexing, retrieval, and searchability, leading to more efficient document organization. In turn, it streamlines downstream tasks such as compliance verification, auditing, and regulatory reporting. Moreover, this methodology contributes to the field of document layout analysis by implicitly capturing layout similarities during clustering. Finally, our research advances understanding of the effectiveness and applicability of different multimodal models in unsupervised document classification tasks, providing valuable information for future work in this area.

To achieve this, we use embeddings from several state-of-the-art pre-trained multimodal models, including SBERT (for text-only baseline) \cite{reimers2019sbert}, LayoutLMv1 \cite{Xu2020}, LayoutLMv3 \cite{Xu2022}, DiT \cite{Li2022}, Donut \cite{Kim2022}, ColPali \cite{Faysse2025}, Gemma3 \cite{gemma32025}, and InternVL3 \cite{internvl32025}. These models have demonstrated remarkable performance in various document understanding tasks by effectively integrating information from different modalities. By evaluating their embeddings for the task of unsupervised document clustering, we aim to provide insights into their strengths and weaknesses in capturing the nuances of document types and templates.

The main contributions of this paper are as follows:
\begin{itemize}
    \item A model-agnostic unsupervised pipeline for clustering documents at category and template granularity, including a token-type--aware projection for encoders that mix text and vision tokens, and a practical {HDBSCAN + $k$-NN} variant to obtain noise-free partitions without labels.
    \item We evaluate the effectiveness of embeddings from several state-of-the-art pre-trained multimodal models (LayoutLMv1, LayoutLMv3, DiT, Donut, ColPali, Gemma3, and InternVL3) for this task, along with a text-only baseline (SBERT).
    \item We demonstrate the potential of multimodal embeddings to distinguish between different templates within the same document category, a capability often lacking in traditional text-based methods.
    \item We provide a comprehensive analysis of the advantages and disadvantages of different multimodal models for unsupervised document clustering, offering valuable insights for practitioners and researchers in IDP, document layout analysis, and unsupervised document classification.
\end{itemize}

The remainder of this paper is structured as follows. Section~\ref{sec:related_work} reviews prior research on deep learning-based document clustering, multi-view document clustering, and multimodal language models, situating our contribution within the existing literature. Section~\ref{sec:methodology} details the proposed methodology: it explains how last-layer hidden states are projected into fixed-size embeddings, describes the eight pre-trained encoders we evaluate, and outlines the algorithms used for clustering. Section~\ref{sec:results_discussion} presents the empirical study. Section~\ref{subsec:hyperparameter_selection} describes the procedure adopted for hyperparameter selection. Section~\ref{subsec:evaluation_metrics} formalizes the external and internal evaluation metrics. Sections~\ref{subsec:template_clustering_results} and~\ref{subsec:document_clustering_results} report template- and document-level clustering results on five heterogeneous corpora. Section~\ref{subsec:ablation_studies} analyzes the impact of pooling strategy and model scale. Section~\ref{subsec:limitations} examines generalization to multi-page documents, situates our approach relative to weakly and semi-supervised alternatives, details post-clustering consolidation, and evaluates robustness to OCR noise and multilingual inputs. Finally, Section~\ref{sec:conclusions} synthesizes the findings.

\section{Related Work}
\label{sec:related_work}

\paragraph{Deep Learning-Based Document Clustering.}

Deep learning has advanced document clustering by learning representations rather than bag-of-words. Contractive autoencoders and deep embedding clustering improve robustness and clusterability \cite{Diallo2021,Xie2016}. Additional lines of work address short-text representation learning \cite{RepresentationLearningShortTextClustering}, regularized concept factorization \cite{Yan2017}, and attention-weighted BERT for term importance \cite{TextClusteringWeightedBERT}. Pre-trained language models such as BERT provide contextualized embeddings that are aggregated into document vectors \cite{Devlin2019,Liu2019}. More recent studies employ LLM embeddings for clustering across domains \cite{PETUKHOVA2025100} and use LLM-generated summaries as interpretable centroids \cite{kLLMmeans2025}. Multimodal and multitask pretraining for document representations has also been explored \cite{TowardsMultimodalMultitask2022}. In contrast, we evaluate off-the-shelf multimodal encoders for unsupervised document and template clustering, with a focus on layout and visual cues in addition to text.

\paragraph{Multi-View Document Clustering.}

Multi-view clustering exploits complementary representations to improve grouping \cite{Fang2023}. For documents, recent methods include a multi-structure processor with hybrid ensemble clustering \cite{Bai2024}, deep subspace approaches that learn shared latent spaces \cite{Zhu2024,Brbic2018}, joint contrastive triple-learning \cite{Hu2023}, enhanced semantic embedding and hierarchical consensus learning for deep multi-view clustering \cite{Bai2021,Bai2024b}, robust ensemble modeling that preserves high-order correlations \cite{Zhao2023}, and shared generative latent representations \cite{Yin2020}. These techniques typically fuse alternative textual views or algorithmic representations rather than distinct modalities. In contrast, multimodal language models learn joint representations over text, layout, and visual signals. Using such multimodal embeddings for document clustering is a distinct and comparatively underexplored direction that we investigate in this work.

\paragraph{Multimodal Language Models.}

Multimodal language models now jointly encode text, layout, and vision for document understanding. The LayoutLM family integrates layout in v1 and adds visual signals with expanded pretraining in v2, and v3 introduces a unified masked-language and masked-image objective \cite{Xu2020,Xu2021,Xu2022}. Vision-focused models advance document image understanding and OCR-free processing, including DiT \cite{Li2022} and Donut \cite{Kim2022}. ColPali produces multi-vector embeddings for retrieval \cite{Faysse2025}. General-purpose vision–language models such as Gemma3 \cite{gemma32025} and InternVL3 \cite{internvl32025} extend these capabilities with scalable pretraining and strong zero-shot performance. Additional architectures combine text, image, and spatial cues. TILT learns from all three modalities \cite{Powalski2021}. TrOCR targets transformer-based OCR \cite{Li2023}. DocFormer fuses text, vision, and spatial features \cite{Appalaraju2021}. DocBERT adapts BERT for document classification \cite{Adhikari2019}. LiGT explores layout-infused generative reasoning \cite{LiGT2025}. Kosmos-2.5 parses text-intensive images \cite{Lv2023}. UDOP and Unified-IO pursue broad multimodal processing \cite{Tang2023,Lu2022}. HRVDA scales to high-resolution documents \cite{Liu2024}. Despite these advances, the suitability of multimodal embeddings for unsupervised document clustering, especially for discovering templates within a category, remains underexplored. We therefore evaluate the clustering performance of embeddings from several of these models for both document-type and template-level organization.

\section{Methodology}
\label{sec:methodology}

\begin{figure}[ht]
     \centering
     \includegraphics[width=0.9\columnwidth]{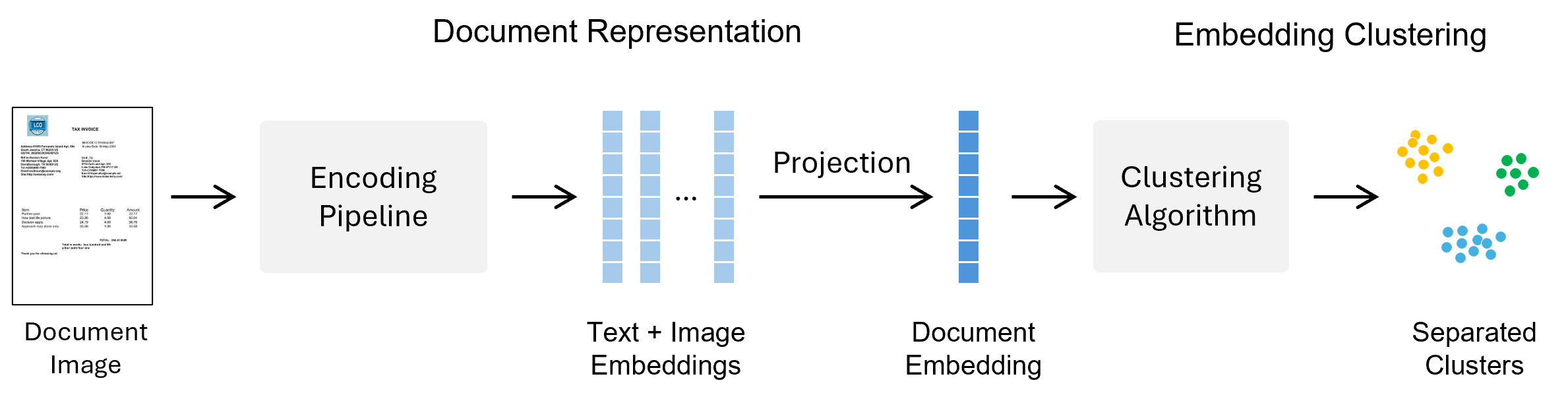}
     \caption{Clustering approach through multimodal embeddings.}
     \label{fig:clustering_pipeline}
\end{figure}

This section presents our clustering approach in detail using different types of textual and image embeddings. It also explains the methodology adopted to evaluate our method. As illustrated in Figure \ref{fig:clustering_pipeline}, the core of the approach involves extracting fixed-size document-level representations from a transformer-based model encoder and clustering these representations using classical unsupervised algorithms. We first explain the projection strategies used to transform the models' last hidden states into single vectors suitable for clustering. Then, we describe the selected embedding models and the clustering algorithms.

\subsection{Document Representation via Last Hidden State Projections}

Let $\mathbf{H} \in \mathbb{R}^{L \times D}$ denote the last hidden state tensor output from the encoder of a transformer model for a single document, where $L$ is the sequence length and $D$ is the hidden dimension. To enable document-level clustering, we must reduce this 2D tensor to a single vector $\mathbf{v} \in \mathbb{R}^d$ (typically $d = D$ or smaller).

A simple but effective strategy is to compute the mean across the sequence length: 
\begin{equation} \mathbf{v} = \frac{1}{L} \sum_{i=1}^{L} \mathbf{H}[i, :]. 
\end{equation} 
This approach performs well when all token embeddings in $\mathbf{H}$ belong to a homogeneous feature space. For instance, Donut's encoder outputs image patch embeddings exclusively, making mean pooling a reliable choice.

However, models such as LayoutLMv3 produce a mix of textual and visual embeddings. In such cases, mean pooling over all tokens is inappropriate, as it blends heterogeneous features. Instead, we employ a hybrid strategy. Let $\mathbf{H}_t \in \mathbb{R}^{L_t \times D}$ and $\mathbf{H}_v \in \mathbb{R}^{L_v \times D}$ represent the text and image embeddings, respectively, such that $L = L_t + L_v$ and $\mathbf{H} = [\mathbf{H}_t; \mathbf{H}_v]$. Our approach proceeds as follows:

\begin{enumerate} 
\item Compute the mean of the text embeddings: \begin{equation} 
\mathbf{v}_t = \frac{1}{L_t} \sum_{i=1}^{L_t} \mathbf{H}_t[i, :]. 
\end{equation} 
\item To reduce the spatial complexity of the image embeddings, apply 1D max-pooling with a kernel size of $k$  along the feature dimension: \begin{equation} 
\bar{\mathbf{H}}_v = \text{MaxPool}_{k}(\mathbf{H}_v), 
\end{equation}
where $\bar{\mathbf{H}}_v \in \mathbb{R}^{L_v \times N}$, with $N < D$.
\item Compute the mean of the pooled image embeddings: \begin{equation} 
\mathbf{v}_v = \frac{1}{L_v} \sum_{i=1}^{L_v} \bar{\mathbf{H}}_v[i, :]. 
\end{equation} 
\item Concatenate text and pooled image embeddings: \begin{equation} 
\mathbf{v} = [\mathbf{v}_t; \mathbf{v}_v]. 
\end{equation} 
\end{enumerate}

This strategy preserves key textual and visual cues without excessively increasing dimensionality. Alternatively, dimensionality reduction can be achieved by principal component analysis (PCA) \cite{pearson1901pca} on $\mathbf{H}_t$ and $\mathbf{H}_v$. Finally, an autoencoder can also be trained to reconstruct the concatenated representations $[\mathbf{H}_t; \mathbf{H}_v]$, using its bottleneck representation as the document embedding. Although the latter approach delegates the optimal strategy for projecting the final hidden states to the trained model itself, it typically demands substantial amounts of training data and considerable computational resources.

\subsection{Embedding Models}

The selection of pre-trained models is critical for effectively capturing the multimodal characteristics inherent in documents. Each model contributes uniquely by addressing distinct dimensions of document information. Sentence-BERT (SBERT) \cite{reimers2019sbert} serves as a text-only baseline; it represents a specialized variant of BERT designed explicitly to generate semantically meaningful sentence embeddings. Using SBERT allows one to evaluate the incremental benefit provided by incorporating layout and visual modalities into clustering tasks. 

LayoutLMv1 \cite{Xu2020} introduces layout awareness by augmenting BERT with two-dimensional positional embeddings, effectively integrating textual and spatial layout information. Its inclusion enables an assessment of the specific contribution of layout features relative to text alone. LayoutLMv3 \cite{Xu2022} further enriches this integration by incorporating visual features into its architecture, employing a unified framework for masking text and images. This facilitates learning cross-modal interactions among textual content, layout structure, and visual elements, highlighting the additional value of visual information. 

The Document Image Transformer (DiT) \cite{Li2022}, on the other hand, primarily targets visual aspects. DiT, a vision transformer pre-trained through self-supervised learning on extensive document image datasets, allows the evaluation of the efficacy of visual representations independently of textual and layout contexts. Donut \cite{Kim2022}, an OCR-free Document Understanding Transformer, processes images directly without relying on external OCR modules. Using transformer-based architectures to generate textual content directly from visual inputs, it provides insights into the performance of end-to-end models. 

ColPali \cite{Faysse2025}, a vision-language model tailored for efficient document retrieval, is trained to generate robust multi-vector embeddings that encapsulate both visual and textual semantics. Incorporating ColPali allows exploration of embeddings explicitly optimized for retrieval tasks within the clustering context. 

Gemma3 \cite{gemma32025} is a multimodal vision–language model family that couples a language backbone with a SigLIP \cite{siglip2023} visual encoder and a projection layer to align visual and language representations. In this work, we use the 4B variant: we take the image features after their projection into the language space and apply mean pooling to obtain a single embedding for clustering.

Among recent vision–language models, InternVL3 \cite{internvl32025} represents a family that pairs a Qwen2.5 language backbone \cite{qwen2025} with an InternViT visual encoder. In this work, we use the 14B variant, which couples Qwen2.5-14B with InternViT-300M. It is trained through a native multimodal pre-training paradigm, where the model is exposed to both text-only corpora and multimodal data simultaneously during one unified pre-training stage, rather than adapting a text-only LLM in a secondary phase. For clustering, we apply mean pooling over the projected image features derived from InternViT.

\begin{table}[ht]
\centering
\caption{Overview of the models considered in this study, the input modalities exploited during pre-training, their parameter counts, and their encoder hidden sizes.}
\label{tab:model_overview}
\begin{tabular}{llcc}
\toprule
\textbf{Model} & \textbf{Modalities (pre-training)} & \textbf{No.\ parameters} & \textbf{Hidden size} \\
\midrule
SBERT (Base)      & Text                 & 110M & 384 \\
LayoutLMv1 (Base) & Text, Layout         & 113M & 768 \\
LayoutLMv3 (Base) & Text, Layout, Image  & 113M & 768 \\
DiT (Base)        & Image                & 87M  & 768 \\
Donut (Base)      & Text, Image          & 176M & 768 \\
ColPali           & Text, Image          & 3B   & 128 \\
Gemma3            & Text, Image          & 4B   & 2560 \\
InternVL3         & Text, Image, Video   & 14B  & 1024 \\
\bottomrule
\end{tabular}
\end{table}

\begin{figure}[ht]
     \centering
     \includegraphics[width=0.7\columnwidth]{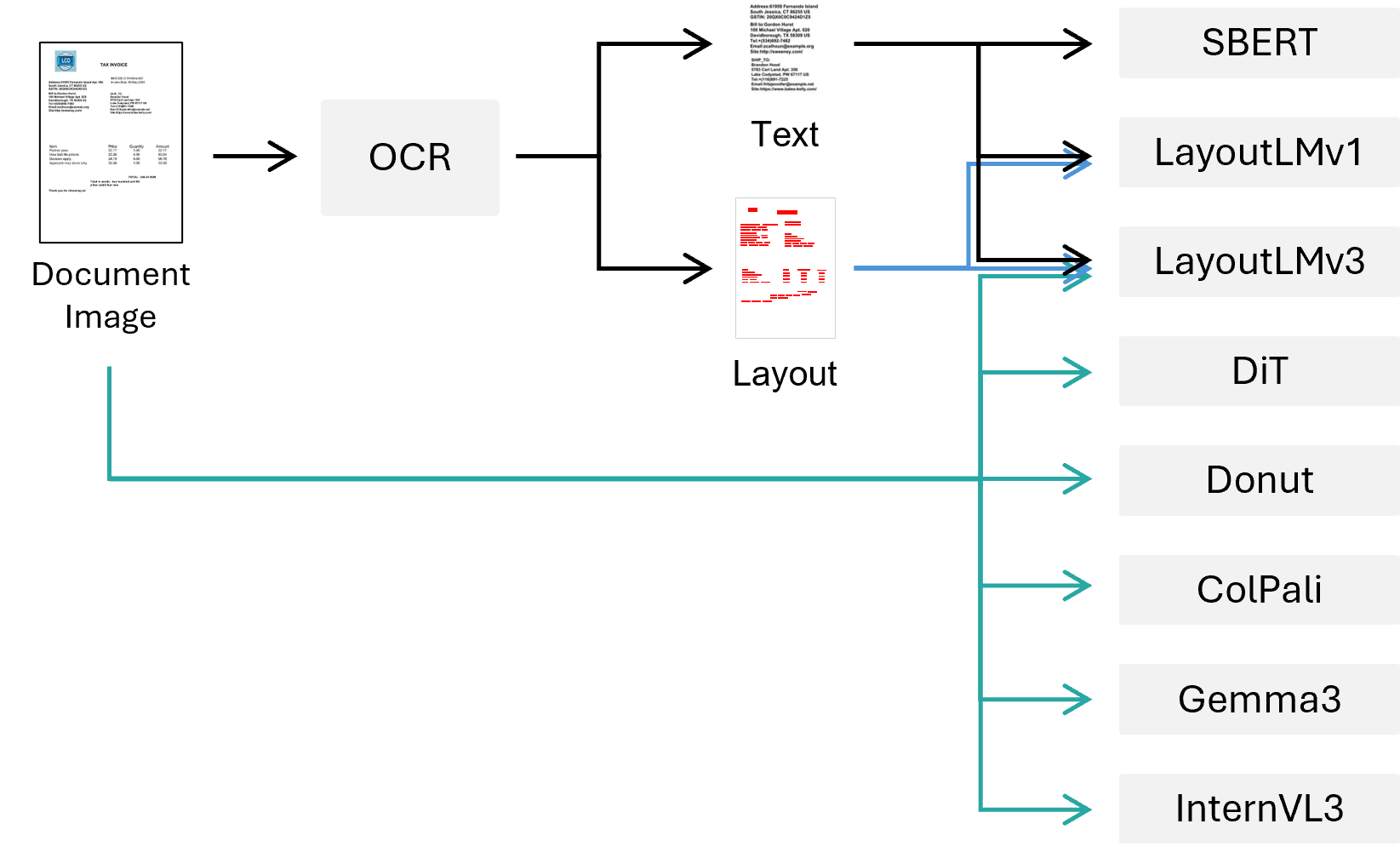}
     \caption{Document processing pipeline of each pre-trained model during inference.}
     \label{fig:different_modalities}
\end{figure}

Table \ref{tab:model_overview} summarizes the eight encoders considered in this study, detailing their pre-training modalities, transformer hidden sizes, and parameter counts. Figure \ref{fig:different_modalities} illustrates how each pre-trained model processes the document image during inference. Through this diverse and carefully curated selection of models, our objective is to thoroughly analyze how different modalities and architectures influence the effectiveness of document clustering according to document types and templates.

\subsection{Clustering Algorithms}

After obtaining the document embeddings from each of the selected models, we employ widely used clustering algorithms as described next.

\subsubsection{$k$-Means}

$k$-Means \cite{forgy1965, lloyd1982, ahmed2020} is a centroid-based partitioning algorithm that aims to divide $n$ samples into $k$ clusters by iteratively assigning each sample to the cluster with the nearest mean (centroid). The algorithm proceeds as follows:

\begin{enumerate}
    \item Initialize $k$ centroids randomly or using a specific initialization method (e.g. $k$-Means++).
    \item Assign each data point to the cluster whose centroid is closest (using a distance metric such as the Euclidean distance).
    \item Recalculate the centroids of each cluster as the mean of all data points assigned to that cluster.
    \item Repeat steps 2 and 3 until the cluster assignments no longer change or a maximum number of iterations is reached.
\end{enumerate}

$k$-Means is relatively simple to implement and computationally efficient, making it suitable for large datasets. However, it requires specifying the number of clusters ($k$) beforehand, which may not always be known in unsupervised settings. It also assumes that clusters are spherical and of similar size and that their performance can be sensitive to the initial choice of centroids.

\subsubsection{DBSCAN}

DBSCAN (Density-Based Spatial Clustering of Applications with Noise) \cite{ester1996} is a density-based clustering algorithm that groups together data points that are closely packed together, marking as outliers points that lie alone in low-density regions. DBSCAN defines clusters based on the density of data points in the feature space. It relies on two parameters:

\begin{itemize}
    \item \textbf{$\epsilon$:} The maximum distance between two data points for them to be considered neighbors.
    \item \textbf{\texttt{minPts}:} The minimum number of data points required to form a dense region (core point).
\end{itemize}

The algorithm works by identifying core points (points with at least \texttt{minPts} neighbors within distance $\epsilon$), border points (points that are neighbors of a core point but are not core points themselves) and noise points (points that are neither core nor border points). The clusters are then formed by connecting core points that are within $\epsilon$ distance of each other, along with their border points.

The key advantages of DBSCAN are that it does not require specifying the number of clusters in advance and that it can discover clusters of arbitrary shapes. It is also robust to outliers. However, DBSCAN can struggle with clusters of varying densities, and its performance is sensitive to the choice of parameters $\epsilon$ and \texttt{minPts}.

\subsubsection{HDBSCAN + $k$-NN}

HDBSCAN \cite{campello2013} extends DBSCAN by building a hierarchy of clusters over a range of density thresholds and then extracting a flat clustering based on cluster stability. In practice, HDBSCAN may return a substantial number of points labeled as noise. To mitigate this, we adopt a two-stage procedure that combines HDBSCAN with a post hoc assignment of $k$-nearest neighbors ($k$-NN) \cite{FixHodges1951}.

\begin{enumerate}
\item Run HDBSCAN with a chosen metric and hyperparameters (e.g., \texttt{min\_cluster\_size}, \texttt{min\_samples}) to obtain initial cluster labels, where noise points are assigned label $-1$.
\item Train a $k$-NN classifier on the \emph{non-noise} points (labels $\neq -1$), using their HDBSCAN labels as targets.
\item Predict cluster labels for all points labeled as noise by HDBSCAN using the trained $k$-NN model.
\end{enumerate}

This hybrid approach preserves HDBSCAN's ability to discover clusters without specifying $k$ while ensuring that, after the $k$-NN reassignment, no data points remain unlabeled as noise. In settings where HDBSCAN provides soft cluster membership probabilities, these can optionally be used as class weights during $k$-NN training.

\subsubsection{BIRCH}

BIRCH (Balanced Iterative Reducing and Clustering using Hierarchies) \cite{birch1996} incrementally builds a compact summary of the dataset called a Clustering Feature (CF) tree. Each node stores sufficient statistics (number of points, linear sum, and squared sum) that enable efficient updates and distance computations.

\begin{enumerate}
\item \textbf{CF-tree construction:} Insert data points one by one into the CF tree using a specified distance threshold $T$ and branching factor $B$, merging or splitting leaf entries to maintain compact summaries.
\item \textbf{Global clustering (optional):} Apply a global clustering algorithm (e.g., $k$-Means) to the leaf entries (subcluster centroids) to obtain the final partition; alternatively, the leaf entries themselves can serve as the final clustering.
\end{enumerate}

BIRCH is well-suited to large or streaming datasets due to its single-pass, memory-efficient construction and its ability to reduce data while preserving cluster structure. Its limitations include a preference for roughly spherical clusters and sensitivity to the threshold parameter $T$, which controls leaf entry compactness.

\section{Results and Discussion}
\label{sec:results_discussion}

This section comprises six parts. First, we describe the procedure adopted for hyperparameter selection (\S\ref{subsec:hyperparameter_selection}), detailing how the clustering algorithms were tuned and, in the case of $k$-Means, how initialization was handled. We then formalize the external and internal evaluation metrics that will be used throughout (\S\ref{subsec:evaluation_metrics}). Section \ref{subsec:template_clustering_results} presents the template-level clustering experiments, whose objective is to group pages that instantiate an identical layout within the same document category; the datasets specific to this task are introduced at the beginning of this subsection. Section \ref{subsec:document_clustering_results} addresses document-level clustering, where each single-page document is clustered according to its general type, and the corresponding corpora are likewise detailed in situ. Section \ref{subsec:ablation_studies} reports ablation studies that isolate architectural capacity and pooling strategy: we contrast the base and large-scale LayoutLMv1 checkpoints and compare the mean pooling of the final hidden states with the conventional \texttt{[CLS]} token. Finally, Section~\ref{subsec:limitations} discusses generalization beyond single-page templates, positioning with respect to weakly and semi-supervised alternatives, post-clustering consolidation, and robustness to OCR noise and multilingual inputs. 

The OCR model used to extract textual content from the document images is the CRNN model \cite{Shi2017}, chosen for its robustness across various document layouts and its efficiency. Apart from a light deskew applied only when obvious rotation was detected, no preprocessing, normalization or denoising is performed. Note that $k$-Means and BIRCH are supplied with the ground-truth number of clusters, giving it a distinct advantage over DBSCAN and HDBSCAN + $k$-NN. In all experiments $k$-Means is initialized with the greedy $k$-Means++ scheme.

\subsection{Hyperparameter Selection}
\label{subsec:hyperparameter_selection}

For clustering algorithms that do not require the number of clusters as input, a robust hyperparameter selection process was essential to ensure optimal performance. To this end, a systematic grid search was implemented to tune the relevant parameters, with the objective of maximizing the internal validation metric, namely the silhouette score. This ensured that the parameterization adhered to a strictly unsupervised methodology and reflected the inherent structure of the data. For DBSCAN, the grid search explored a range of values for both $\epsilon$ and \texttt{minPts}, while for HDBSCAN the search concentrated on different values of the \texttt{min\_cluster\_size} and \texttt{min\_samples} parameters, which largely governs the resulting cluster structure. In the case of BIRCH, a grid search was likewise carried out on the \textit{threshold} parameter, which controls the compactness of the subclusters in the CF-tree. However, unlike DBSCAN and HDBSCAN, the final evaluation of BIRCH was performed in an oracle mode, in which the ground-truth number of clusters was supplied during the global clustering step. This evaluation setup was aligned with that of $k$-Means, which was also assessed exclusively under an oracle condition. In all experiments, $k$-Means was initialized with the greedy $k$-Means++ scheme to improve convergence and stability.

\subsection{Evaluation Metrics}\label{subsec:evaluation_metrics}

To quantitatively assess the effectiveness of our clustering approach, we employ a comprehensive set of metrics computed using \texttt{Scikit-learn} \cite{pedregosa2011scikit}. The chosen metrics offer insight into different aspects of cluster quality, capturing both external agreement with known labels and internal structural properties of the clusters. Specifically, the adjusted rand index (ARI) \cite{steinley2004properties} measures the similarity between predicted clusters and ground truth classes, adjusting for chance agreement, thus providing an unbiased external clustering evaluation. Similarly, normalized mutual information (NMI) \cite{vinh2009information} quantifies the information shared between predicted and actual labels, also adjusted for randomness, highlighting the coherence of the clustering results.

Furthermore, we use the homogeneity score (HS) and the completeness score (CS), both introduced by Rosenberg and Hirschberg \cite{rosenberg2007v}. HS assesses the extent to which each cluster contains only members of a single class, while CS assesses whether all members of a given class are assigned to the same cluster.

In addition, internal cluster quality is evaluated using the silhouette score (SS) \cite{rousseeuw1987silhouettes}, which measures compactness within clusters and their separation from other clusters based solely on distance metrics, independent of any ground truth labels. The number of predicted clusters (PC) is also reported to understand the granularity of the clustering and its impact on the metrics. Complementarily, we record the percentage of noise points that DBSCAN designates as outliers, providing a direct indication of how much data each representation fails to accommodate within any cluster. Collectively, these metrics allow for a nuanced and comprehensive evaluation of our clustering methods in varying experimental scenarios.

\subsection{Template-Level Clustering}
\label{subsec:template_clustering_results}

\subsubsection{Datasets}

This scenario focuses on discerning the underlying layout templates within documents of the same category. Three datasets were used to evaluate the performance of each embedding model on this task. Table \ref{tab:template_clustering} shows the characteristics of each dataset considered. The FATURA \cite{limam2025fatura} dataset is a synthetic invoice dataset comprising 10,000 images generated from 50 distinct templates. Each template is designed to mimic real-world invoice layouts, incorporating various textual and structural elements. The dataset includes detailed annotations in multiple formats, facilitating tasks such as document layout analysis and key-value extraction. Its synthetic nature ensures a wide variety of layouts while mitigating privacy concerns associated with real invoices. We then extended the FATURA dataset by introducing significant visual perturbations, including noise, shadows, folds, and ink artifacts. These transformations simulate real-world degradations commonly encountered in scanned or photographed documents, providing a more challenging benchmark for evaluating the robustness of document clustering algorithms. The resulting dataset is denoted here Transformed FATURA.

We also added an internal real-world dataset of Mexican death certificates characterized by moderate noise levels. These documents present a consistent template structure, making them suitable for assessing the effectiveness of clustering methods in identifying layout variations within a document type.

\begin{table}[ht]
\centering
\caption{Datasets for template-level clustering.}
\begin{tabular}{lccc}
\hline
\textbf{Dataset} & \textbf{Size} & \textbf{No. classes} & \textbf{Ref.} \\
\hline
FATURA & 10,000 & 50 & \cite{limam2025fatura} \\
Transformed FATURA & 10,000 & 50 & Internal \\
Mexican death certificates & 250 & 9 & Internal \\
\hline
\end{tabular}
\label{tab:template_clustering}
\end{table}

\subsubsection{Experimental Results}

The three corpora enable a controlled inspection of how text-only, layout-aware, vision-only, and vision-language representations shape the topology of template-level embeddings, and consequently, the behavior of each clustering algorithm.

% ============================================================
% FATURA – template-level clustering results
% ============================================================
\begin{table}[!ht]
\caption{Template-level clustering results on the clean FATURA synthetic-invoice corpus. Metrics are adjusted rand index (ARI), normalized mutual information (NMI), homogeneity score (HS), completeness score (CS), silhouette score (SS), the number of predicted clusters (PC) and the percentage of noise points returned by DBSCAN. Ground truth has 50 clusters. The top-performing scores are shown in boldface.}
\centering
\small
\begin{tabular}{llccccccr}
\hline
Model & Clustering Alg. & ARI & NMI & HS & CS & SS & PC & \% Noise \\
\hline
SBERT        & $k$-Means      & 0.7615 & 0.9306 & 0.918  & 0.9436 & 0.2584 & --  & -- \\
SBERT        & DBSCAN         & 0.0119 & 0.2529 & 0.152  & 0.7527 & 0.0286 & 9   & 6.9 \\
SBERT        & HDBSCAN + $k$-NN  & 0.8634 & 0.9490 & 0.9774 & 0.9223 & 0.2646 & 75  & -- \\
SBERT        & BIRCH          & 0.8551 & 0.9462 & 0.9418 & 0.9507 & 0.2719 & --  & -- \\
\hline
LayoutLMv1   & $k$-Means      & 0.8576 & 0.9604 & 0.9499 & 0.9710 & 0.2763 & --  & -- \\
LayoutLMv1   & DBSCAN         & 0.1380 & 0.7029 & 0.5741 & 0.9063 & 0.0124 & 30  & 15.92  \\
LayoutLMv1   & HDBSCAN + $k$-NN  & 0.8829 & 0.9703 & 0.9601 & 0.9806 & 0.2932 & \textbf{49} & -- \\
LayoutLMv1   & BIRCH          & 0.9996 & 0.9997 & 0.9997 & 0.9997 & 0.2983 & --  & -- \\
\hline
LayoutLMv3   & $k$-Means      & 0.7422 & 0.9091 & 0.9010 & 0.9174 & 0.6007 & --  & -- \\
LayoutLMv3   & DBSCAN         & 0.8579 & 0.9479 & 0.9960 & 0.9042 & 0.6811 & 94  & 0.59   \\
LayoutLMv3   & HDBSCAN + $k$-NN  & 0.8334 & 0.9429 & 0.9998 & 0.8921 & 0.6757 & 104 & -- \\
LayoutLMv3   & BIRCH          & 0.7563 & 0.9224 & 0.9123 & 0.9328 & 0.5479 & --  & -- \\
\hline
DiT          & $k$-Means      & 0.9644 & 0.9882 & 0.9872 & 0.9893 & 0.6503 & --  & -- \\
DiT          & DBSCAN         & 0.9179 & 0.9667 & \textbf{1} & 0.9356 & \textbf{0.7375} & 79  & \textbf{0} \\
DiT          & HDBSCAN + $k$-NN  & 0.7367 & 0.9052 & \textbf{1} & 0.8269 & 0.7305 & 164 & -- \\
DiT          & BIRCH          & 0.9923 & 0.9973 & 0.9971 & 0.9974 & 0.6608 & --  & -- \\
\hline
Donut        & $k$-Means      & 0.9708 & 0.9921 & 0.9907 & 0.9935 & 0.6937 & --  & -- \\
Donut        & DBSCAN         & 0.9240 & 0.9856 & 0.9717 & \textbf{1} & 0.6756 & 46  & \textbf{0} \\
Donut        & HDBSCAN + $k$-NN  & 0.9385 & 0.9742 & \textbf{1} & 0.9496 & 0.6860 & 75  & -- \\
Donut        & BIRCH          & 0.9759 & 0.9950 & 0.9929 & 0.9970 & 0.6887 & --  & -- \\
\hline
ColPali      & $k$-Means      & 0.7862 & 0.9241 & 0.9151 & 0.9334 & 0.1698 & --  & -- \\
ColPali      & DBSCAN         & 0.0050 & 0.0788 & 0.0437 & 0.3984 & 0.1176 & 2   & 15.38 \\
ColPali      & HDBSCAN + $k$-NN  & 0.3696 & 0.7202 & 0.6616 & 0.7902 & 0.1283 & 35  & -- \\
ColPali      & BIRCH          & 0.7735 & 0.9004 & 0.8945 & 0.9062 & 0.1744 & --  & -- \\
\hline
Gemma3       & $k$-Means      & 0.8496 & 0.9617 & 0.9527 & 0.9708 & 0.4234 & --  & -- \\
Gemma3       & DBSCAN         & 0.9388 & 0.9718 & 0.9886 & 0.9556 & 0.4331 & 71  & 2.14 \\
Gemma3       & HDBSCAN + $k$-NN  & 0.9812 & 0.9924 & \textbf{1} & 0.9849 & 0.4657 & 56  & -- \\
Gemma3       & BIRCH          & \textbf{1} & \textbf{1} & \textbf{1} & \textbf{1} & 0.4707 & --  & -- \\
\hline
InternVL3    & $k$-Means      & 0.9261 & 0.9790 & 0.9756 & 0.9824 & 0.5308 & --  & -- \\
InternVL3    & DBSCAN         & 0.9310 & 0.9786 & 0.9922 & 0.9653 & 0.5502 & 61  & 0.24 \\
InternVL3    & HDBSCAN + $k$-NN  & 0.9510 & 0.9826 & \textbf{1} & 0.9659 & 0.5526 & 62  & -- \\
InternVL3    & BIRCH          & 0.9518 & 0.9897 & 0.9858 & 0.9937 & 0.5533 & --  & -- \\
\hline
\end{tabular}
\label{tab:template_clustering_fatura}
\end{table}

\paragraph{FATURA.} Table \ref{tab:template_clustering_fatura} reports the complete set of template-level clustering scores obtained on the clean FATURA corpus. It shows that the raster signal coincides almost perfectly with the template identity, and the purely visual encoders are virtually indistinguishable from a supervised classifier. ARIs are frequently very high, and multiple combinations have a near-perfect result or are perfect. The standout is Gemma3 + BIRCH, which achieves perfect agreement across all external metrics, with $k = 50$ provided. DiT and Donut achieve ARI = 0.9923 / 0.9759, NMI = 0.9973 / 0.995, HS = 0.9971 / 0.9929, and CS = 0.9974 / 0.997 with BIRCH, which is superior to $k$-Means. When DBSCAN must infer the number of clusters, these models still exceed 0.90 ARI and predict 46–79 clusters while declaring 0\% noise, indicating that visual features form compact, well-separated regions in the embedding space. LayoutLMv3, which fuses lexical, positional, and visual cues, follows at ARI = 0.7422. SBERT performs decently but behind multimodal models: best ARI = 0.8634 with HDBSCAN + $k$-NN, while $k$-Means lags at 0.7615 ARI. Its DBSCAN run collapses to 9 clusters, ARI = 0.0119, and 6.9\% noise. ColPali, a vision-language retriever that should, in principle, exploit the raster image, behaves unexpectedly as SBERT: its $k$-Means ARI is 0.7862 and its DBSCAN ARI drops below 0.02. When the value of $k$ must be inferred, HDBSCAN combined with $k$-NN significantly outperforms DBSCAN, achieving a homogeneity score of 0.6616 and a completeness score of 0.7902.

Algorithmically, two patterns stand out. First, BIRCH and $k$-Means, when provided with $k$ = 50, consistently rank among the best performers -- frequently achieving ARI scores greater than 0.85 and occasionally reaching perfect scores, as seen with Gemma3. Second, HDBSCAN + $k$-NN almost always improves over plain DBSCAN on ARI and NMI, while driving the percentage of noise points to 0\% by design (since $k$-NN assigns outliers back to clusters). Where DBSCAN discards between roughly 0.2\% and 2.1\% of points as noise for the best performers (e.g., InternVL3 DBSCAN = 0.24\%, Gemma3 DBSCAN = 2.14\%), HDBSCAN + $k$-NN neutralizes that loss without decreasing ARI or NMI; in fact it often slightly improves them (e.g., SBERT from ARI 0.0119 with DBSCAN to 0.8634 with HDBSCAN + $k$-NN).

% ============================================================
% Transformed FATURA (augmented invoices) – template-level clustering
% ============================================================
\begin{table}[ht]
\caption{Template-level clustering on Transformed FATURA. Metrics are adjusted rand index (ARI), normalized mutual information (NMI), homogeneity score (HS), completeness score (CS), silhouette score (SS), the number of predicted clusters (PC) and the percentage of noise points returned by DBSCAN. Ground truth has 50 clusters. The top-performing scores are shown in boldface.}
\centering
\small
\begin{tabular}{llccccccr}
\hline
Model & Clustering Alg. & ARI & NMI & HS & CS & SS & PC & \% Noise \\
\hline
SBERT        & $k$-Means        & \textbf{0.7378} & 0.8642 & 0.8571 & 0.8715 & 0.1437 & -- & -- \\
SBERT        & DBSCAN           & 0.0043 & 0.0791 & 0.0446 & 0.3464 & 0.1183 & 4  & 13.85 \\
SBERT        & HDBSCAN + $k$-NN    & 0      & 0.0011 & 0.0006 & 0.1356 & 0.5362 & 2  & -- \\
SBERT        & BIRCH            & 0.6878 & 0.8510 & 0.8440 & 0.8581 & 0.1322 & -- & -- \\
\hline
LayoutLMv1   & $k$-Means        & 0.3574 & 0.6527 & 0.6443 & 0.6613 & 0.0694 & -- & -- \\
LayoutLMv1   & DBSCAN           & 0.0030 & 0.0435 & 0.0243 & 0.2114 & 0.2648 & 3  & 15.63 \\
LayoutLMv1   & HDBSCAN + $k$-NN    & 0      & 0.0012 & 0.0006 & 0.1327 & \textbf{0.9781} & 2  & -- \\
LayoutLMv1   & BIRCH            & 0.5141 & 0.7806 & 0.7729 & 0.7885 & 0.0590 & -- & -- \\
\hline
LayoutLMv3   & $k$-Means        & 0.4333 & 0.6735 & 0.6658 & 0.6814 & 0.2411 & -- & -- \\
LayoutLMv3   & DBSCAN           & 0.0986 & 0.4862 & 0.3809 & 0.6718 & 0.1360 & 24 & 14.52 \\
LayoutLMv3   & HDBSCAN + $k$-NN    & 0.2092 & 0.7094 & 0.6668 & 0.7579 & 0.3418 & \textbf{66} & -- \\
LayoutLMv3   & BIRCH            & 0.4545 & 0.7205 & 0.7089 & 0.7326 & 0.2289 & -- & -- \\
\hline
DiT          & $k$-Means        & 0.2967 & 0.5261 & 0.5185 & 0.5340 & 0.1759 & -- & --  \\
DiT          & DBSCAN           & 0.0006 & 0.0166 & 0.0091 & 0.0947 & 0.2995 & 4  & 11.07 \\
DiT          & HDBSCAN + $k$-NN    & 0	     & 0.0011 & 0.0005 & 0.1333 & 0.8206 & 2  & -- \\
DiT          & BIRCH            & 0.2849 & 0.5631 & 0.5491 & 0.5778 & 0.1641 & -- & -- \\
\hline
Donut        & $k$-Means        & 0.1556 & 0.3596 & 0.3529 & 0.3666 & 0.1986 & -- & --  \\
Donut        & DBSCAN           & 0.0004 & 0.0073 & 0.0040 & 0.0380 & 0.3060 & 2  & 14.45 \\
Donut        & HDBSCAN + $k$-NN    & 0      & 0.0023 & 0.0012 & 0.1469 & 0.7262 & 3  & -- \\
Donut        & BIRCH            & 0.1791 & 0.4008 & 0.3866 & 0.416  & 0.1412 & -- & -- \\
\hline
ColPali      & $k$-Means        & 0.7176 & \textbf{0.8672} & \textbf{0.8587} & 0.8759 & 0.1242 & -- & --  \\
ColPali      & DBSCAN           & 0.0014 & 0.0357 & 0.0193 & 0.2289 & 0.1818 & 2  & 10.25 \\
ColPali      & HDBSCAN + $k$-NN    & 0      & 0.0010 & 0.0005 & 0.1905 & 0.7245 & 2  & -- \\
ColPali      & BIRCH            & 0.7022 & 0.8448 & 0.8373 & 0.8525 & 0.1155 & -- & -- \\
\hline
Gemma3       & $k$-Means        & 0.1521 & 0.4254 & 0.4190 & 0.4319 & 0.1086 & -- & -- \\
Gemma3       & DBSCAN           & 0.0030 & 0.0728 & 0.0397 & 0.4355 & 0.2817 & 3  & \textbf{5.51}\\
Gemma3       & HDBSCAN + $k$-NN    & 0.0031 & 0.0741 & 0.0387 & \textbf{0.8909} & 0.4007 & 2  & -- \\
Gemma3       & BIRCH            & 0.2295 & 0.5763 & 0.5618 & 0.5916 & 0.0673 & -- & -- \\
\hline
InternVL3    & $k$-Means        & 0.2601 & 0.5263 & 0.5197 & 0.5330 & 0.1457 & -- & -- \\
InternVL3    & DBSCAN           & 0.0029 & 0.0748 & 0.0400 & 0.5612 & 0.5561 & 3  & 3.16 \\
InternVL3    & HDBSCAN + $k$-NN    & 0      & 0.0032 & 0.0016 & 0.4481 & 0.5840 & 2  & -- \\
InternVL3    & BIRCH            & 0.3426 & 0.6356 & 0.6251 & 0.6465 & 0.1352 & -- & -- \\
\hline
\end{tabular}
\label{tab:template_clustering_transformed_fatura}
\end{table}

\paragraph{Transformed FATURA.} The ranking reverses on Transformed FATURA, where folds, scanner shadows and speckle blur template boundaries without altering token identities. Table \ref{tab:template_clustering_transformed_fatura} presents the clustering results in Transformed FATURA, illustrating how heavy visual perturbations degrade or reshape performance relative to the clean baseline. SBERT’s $k$-Means ARI decreases by only three points to 0.7378 and its NMI remains 0.8642, confirming lexical robustness. ColPali exhibits the same stability, suggesting the hypothesis that its embeddings are effectively text-dominated in this setting. LayoutLMv3 also remains relatively stable with $k$-Means, exhibiting moderate to high NMI but noticeably lower ARI compared to the top two methods. This reflects ongoing fragmentation and merging effects, which ARI is more sensitive to and penalizes more strongly. In contrast, DiT and Donut suffer a sharp decline in all metrics.

DBSCAN now predicts between 2 and 24 clusters for every encoder. In addition, it never exceeds 0.099 ARI, and labels up to 15.6\% of pages as noise. The large decrease in completeness and homogeneity is an immediate diagnostic of modality failure: it indicates that vision-centric models under-segment when visual information is unreliable, grouping distinct templates into overly broad clusters. LayoutLMv3 is the only model that consistently maintains moderate to high values across ARI, homogeneity score, and completeness score. InternVL3 performs particularly poorly: its best ARI on Transformed FATURA is just 0.3426 with BIRCH. $k$-Means yields 0.2601, while DBSCAN and HDBSCAN + $k$-NN perform even worse, with ARIs of 0.0029 and 0, respectively.

% ============================================================
% DEATH-CERT (Mexican death certificates) – template-level clustering
% ============================================================
\begin{table}[ht]
\caption{Template-level clustering on the real scanned Mexican death certificates corpus. Metrics are adjusted rand index (ARI), normalized mutual information (NMI), homogeneity score (HS), completeness score (CS), silhouette score (SS), the number of predicted clusters (PC) and the percentage of noise points returned by DBSCAN. Ground truth has 9 clusters. The top-performing scores are shown in boldface.}
\centering
\small
\begin{tabular}{llccccccr}
\hline
Model & Clustering Alg. & ARI & NMI & HS & CS & SS & PC & \% Noise \\
\hline
SBERT      & $k$-Means     & 0.2210 & 0.5297 & 0.7893 & 0.3986 & 0.0926 & -- & -- \\
SBERT      & DBSCAN        & 0.7462 & 0.6810 & 0.7311 & 0.6374 & 0.2240 & \textbf{8} & 14 \\
SBERT      & HDBSCAN + $k$-NN & 0.8593 & 0.7298 & 0.6128 & 0.9023 & 0.2578 & 3  & -- \\
SBERT      & BIRCH         & 0.2423 & 0.5608 & 0.8114 & 0.4285 & 0.1037 & -- & -- \\
\hline
LayoutLMv1 & $k$-Means     & 0.1533 & 0.3972 & 0.5713 & 0.3044 & 0.3669 & -- & -- \\
LayoutLMv1 & DBSCAN        & 0.0191 & 0.1464 & 0.1393 & 0.1541 & 0.5422 & 5  & 6.4 \\
LayoutLMv1 & HDBSCAN + $k$-NN & -0.0567 & 0.0623 & 0.0500 & 0.0827 & \textbf{0.7047} & 2 & -- \\
LayoutLMv1 & BIRCH         & 0.2403 & 0.4430 & 0.6237 & 0.3435 & 0.2986 & -- & -- \\
\hline
LayoutLMv3 & $k$-Means     & 0.2858 & 0.5637 & 0.8136 & 0.4313 & 0.2996 & -- & -- \\
LayoutLMv3 & DBSCAN        & 0.7713 & 0.6995 & 0.7999 & 0.6216 & 0.4493 & 11 & 9.6 \\
LayoutLMv3 & HDBSCAN + $k$-NN & 0.9619 & 0.8272 & 0.7695 & 0.8941 & 0.5096 & 4  & -- \\
LayoutLMv3 & BIRCH         & 0.3916 & 0.5892 & 0.7902 & 0.4697 & 0.3340 & -- & -- \\
\hline
DiT        & $k$-Means     & 0.0409 & 0.1796 & 0.2825 & 0.1317 & 0.2011 & -- & -- \\
DiT        & DBSCAN        & -0.0386 & 0.0513 & 0.0346 & 0.0992 & 0.2772 & 2 & 11.20 \\
DiT        & HDBSCAN + $k$-NN & 0.0686 & 0.0599 & 0.0492 & 0.0765 & 0.2841 & 2  & -- \\
DiT        & BIRCH         & 0.0268 & 0.1535 & 0.2291 & 0.1154 & 0.2276 & -- & -- \\
\hline
Donut      & $k$-Means     & 0.1883 & 0.4793 & 0.7267 & 0.3576 & 0.3238 & -- & -- \\
Donut      & DBSCAN        & 0.9403 & 0.8668 & \textbf{0.9269} & 0.8139 & 0.2532 & 11 & \textbf{4} \\
Donut      & HDBSCAN + $k$-NN & \textbf{0.9749} & \textbf{0.8906} & 0.8452 & \textbf{0.9412} & 0.2739 & 5  & -- \\
Donut      & BIRCH         & 0.1949 & 0.5046 & 0.7595 & 0.3778 & 0.3324 & -- & -- \\
\hline
ColPali    & $k$-Means     & 0.2083 & 0.5300 & 0.8020 & 0.3958 & 0.1215 & -- & -- \\
ColPali    & DBSCAN        & 0.6771 & 0.6218 & 0.5364 & 0.7396 & 0.3440 & 4  & 6.40 \\
ColPali    & HDBSCAN + $k$-NN & 0.6646 & 0.6391 & 0.5036 & 0.8741 & 0.3551 & 3  & -- \\
ColPali    & BIRCH         & 0.4370 & 0.6656 & 0.8909 & 0.5313 & 0.1370 & -- & -- \\
\hline
Gemma3     & $k$-Means     & 0.3164 & 0.5843 & 0.8310 & 0.4505 & 0.3014 & -- & -- \\
Gemma3     & DBSCAN        & 0.9628 & 0.8800 & 0.8738 & 0.8863 & 0.4596 & 6  & 4.80 \\
Gemma3     & HDBSCAN + $k$-NN & 0.8780 & 0.7980 & 0.8016 & 0.7944 & 0.4800 & 5  & -- \\
Gemma3     & BIRCH         & 0.4067 & 0.6486 & 0.8525 & 0.5234 & 0.3174 & -- & -- \\
\hline
InternVL3  & $k$-Means     & 0.3217 & 0.4765 & 0.6629 & 0.3719 & 0.4446 & -- & -- \\
InternVL3  & DBSCAN        & 0.5810 & 0.6267 & 0.7276 & 0.5504 & 0.3846 & 6  & 13.20 \\
InternVL3  & HDBSCAN + $k$-NN & 0.6153 & 0.6053 & 0.6343 & 0.5788 & 0.3843 & 5  & -- \\
InternVL3  & BIRCH         & 0.2352 & 0.5173 & 0.7688 & 0.3898 & 0.3740 & -- & -- \\
\hline
\end{tabular}
\label{tab:template_clustering_death_cert}
\end{table}

\paragraph{Mexican Death Certificates.} Table \ref{tab:template_clustering_death_cert} lists the clustering results in the Mexican death certificate dataset. In the scanned death certificates corpus, the highly repetitive textual content makes the lexical information nearly invariant across forms, restoring the advantage to models that encode spatial and visual structure. The textual clusters produced by SBERT are coarse, with $k$-Means yielding an ARI of 0.221 and a completeness score of 0.3986. Donut attains the best overall performance: HDBSCAN + $k$-NN yields ARI = 0.9749, NMI = 0.8906, homogeneity = 0.8452, and completeness = 0.9412. LayoutLMv3 ranks closely behind, with an ARI of 0.9619 and an NMI of 0.8567. Pure-image DiT, lacking lexical anchoring, collapses to two clusters with $-0.039$ ARI and 11\% noise, showing that visual tokens alone cannot resolve subtle scanner artefacts. ColPali remains below the other vision-aware encoders (DBSCAN ARI = 0.312, NMI = 0.6771), but still better than LayoutLMv1. DBSCAN performs exceptionally well on Gemma3, achieving an ARI of 0.9628 and an NMI of 0.8800, with 6 clusters and only 4.8\% noise. In contrast, InternVL3 delivers only moderate performance.

\paragraph{Implications for Practice and Research.} Together, the results support three claims. (i) When raster patterns are pristine, vision-only encoders nearly solve unsupervised template discovery, positional features without vision (LayoutLMv1) oversegment, and text-only encoders trail. (ii) The same vision embeddings are the most brittle under covariate shift: heavy augmentation reverses the hierarchy, and lexical signals become the primary source of robustness, though at the cost of merging visually distinct templates. (iii) Architectures that fuse all three channels - lexical tokens, spatial coordinates, and image patches — strike the most favorable robustness–accuracy trade-off. Both LayoutLMv3 and Donut outperform the unimodal baselines when neither text nor vision alone suffices.

Methodologically, $k$-Means and BIRCH tend to overstate performance and mask the modality weaknesses that surface as soon as $k$ must be inferred. DiT and Donut lose more than 0.65 ARI in Transformed FATURA when the oracle disappears, while Donut gains 0.78 ARI on Mexican death certificates by switching from $k$-Means to HDBSCAN + $k$-NN. Joint inspection of homogeneity, completeness and the proportion of noise points therefore offers actionable insight that a single aggregate score would obscure, revealing whether an embedding fragments templates or collapses them indistinct, overly broad clusters.

From a practical perspective, the HDBSCAN + $k$-NN hybrid stands out as a strong baseline when labels are unavailable. It consistently matches or exceeds the performance of oracle-$k$ algorithms on two of the three datasets (FATURA and Mexican death certificates), without the constraint of requiring a predefined $k$. Its 0\% final noise rate also makes it particularly suitable for production scenarios where every document must be assigned to a cluster. However, its failure on Transformed FATURA highlights a key limitation: no clustering algorithm can overcome a fundamentally misaligned embedding space. In such cases, model choice becomes the dominant factor. SBERT and ColPali retained substantial NMI and ARI under transformation, while DiT, Donut, InternVL3, and even Gemma3 showed significant performance drops. This underscores the importance of embedding robustness to distribution shifts or noise as a critical selection criterion, independent of the choice of clustering algorithm.

In summary, state-of-the-art image-centric representations can already achieve near perfect clustering on clean data, yet remain vulnerable to realistic degradations. Textual signals confer orthogonal stability, but cannot disambiguate visually induced differences; and genuine resilience emerges only when the three modalities are integrated so that each compensates for the failure modes of the others.

\subsection{Document-Level Clustering}
\label{subsec:document_clustering_results}

\subsubsection{Datasets}

In this scenario, the objective is to group documents into pre-defined categories based on their content and structure. The dataset used for evaluation is constructed by combining samples from multiple sources, with each contributing dataset representing a distinct class of documents, for example, FATURA for invoices and SROIE for receipts. Table~\ref{tab:document_clustering} summarizes the constituent datasets, indicating their individual characteristics and the number of selected images used to compose the final dataset for document-level clustering. The FATURA dataset and the Mexican death certificates dataset are also used in this context to evaluate the capability of clustering algorithms to distinguish between different types of documents.

SROIE (Scanned Receipt OCR and Information Extraction) \cite{huangicdar2019} is a dataset introduced during the ICDAR 2019 competition, comprising 1,000 scanned receipt images annotated for text localization, OCR, and key information extraction tasks. The receipts vary in layout and content, providing a realistic benchmark for document classification and information retrieval systems.

Finally, the Brazilian Identity Document (BID) public dataset \cite{soares2020bid} encompasses 28,800 images of Brazilian identification documents, including front and back faces of three types of documents: National Driver's Licenses (CNH), Natural Persons Registers (CPF), and General Registrations (RG). The dataset is annotated for document classification, text region segmentation, and OCR tasks, providing a comprehensive resource to evaluate document analysis models in the context of identity verification.

We first randomly selected 200 images each from the FATURA, SROIE, and Mexican death certificates datasets. To complement these, we sampled an additional 200 images from each of the three document categories within the BID dataset — namely, RG, CNH, and CPF — resulting in a final dataset comprising 1,200 images and 6 distinct classes. This curated composition enables a comprehensive evaluation of document clustering techniques across a wide spectrum of document complexity, visual noise, and structural diversity.

\begin{table}[ht]
\centering
\caption{Composition of the dataset used for document-level clustering.}
\begin{tabular}{lccc}
\hline
\textbf{Dataset} & \textbf{Size} & \textbf{Ref.} \\
\hline
FATURA & 200 & \cite{limam2025fatura} \\
SROIE & 200 & \cite{huangicdar2019} \\
Mexican death certificates & 200 & Internal \\
Brazilian Identity Document (BID) & 600 & \cite{soares2020bid} \\
\hline
\end{tabular}
\label{tab:document_clustering}
\end{table}

\subsubsection{Experimental Results}

In the mixed-corpus experiment, the global hierarchy differs sharply from the template-level picture. Table \ref{tab:document_clustering_results} summarizes the results of document-level clustering, showing how effectively each model–algorithm pair separates pages by category. 

\paragraph{$k$-Means.} With the true cardinality supplied, ColPali reaches perfect concordance with the ground truth, achieving ARI = 1, NMI = 1, HS = 1, CS = 1, SS = 0.5581, indicating that its manifold is linearly separable at the class scale. The SBERT baseline is nearly perfect as well, ARI = 0.9890, NMI = 0.9837, HS = 0.9837, CS = 0.9837, SS = 0.3964, showing that token identity and language cues alone suffice to distinguish these six categories when $k$ is known. Among multimodal encoders, Donut performs strongly (ARI = 0.9437, NMI = 0.9471, HS = 0.9467, CS = 0.9475, SS = 0.6253), with the highest $k$-Means silhouette among the top performers, reflecting compact and well-separated clusters. Gemma3 clusters well under the oracle with ARI = 0.7853 and NMI = 0.8427, and its silhouette of 0.6596 is the highest among all $k$-Means runs, suggesting especially tight intra-cluster geometry despite moderate ARI. LayoutLMv3 presents similar external agreement with ARI = 0.7881 and NMI = 0.8419. LayoutLMv1 trails these two with ARI = 0.7094, NMI = 0.8530, HS = 0.8248, CS = 0.8833, SS = 0.4763; its high completeness relative to homogeneity hints at occasional under-segmentation across visually similar categories. InternVL3 performs less well overall (ARI = 0.5195, NMI = 0.6631), yet it achieves a notably high silhouette score (SS = 0.6631), suggesting that its clusters are internally coherent despite limited alignment with the ground-truth labels. Purely visual DiT ranks last under oracle $k$ (HS = 0.4973, CS = 0.5079), consistent with the notion that document categories are largely driven by lexical and symbolic differences that vision-only embeddings struggle to separate.

\paragraph{BIRCH.} With the same advantage of the true $k$, BIRCH reproduces the $k$-Means hierarchy with minor differences. ColPali again achieves perfect scores on all external metrics. SBERT slightly improves to ARI = 0.9909. Donut remains strong with NMI = 0.9524, HS = 0.9514 and CS = 0.9535, while Gemma3 edges up relative to its $k$-Means run, obtaining ARI = 0.8154 and NMI = 0.8725. LayoutLMv1 improves modestly, while LayoutLMv3 is worse under BIRCH than under $k$-Means, suggesting that its manifold responds more favorably to centroid partitioning than to BIRCH’s incremental clustering. InternVL3 and DiT increase slightly relative to $k$-Means, but both remain far from the top tier. Under oracle conditions, the overall picture is therefore stable: language-dominated or highly aligned vision–language embeddings cluster categories almost perfectly; visual-only signals are insufficient; and BIRCH gives broadly comparable results to $k$-Means, with small model-dependent shifts.

% ============================================================
% Mixed-corpus document-level clustering results (FATURA, SROIE,
% Mexican death certificates, BID-3 categories)
% ============================================================
\begin{table}[h!]
\centering
\caption{Document-level clustering results on the six-class mixed corpus. Metrics are adjusted rand index (ARI), normalized mutual information (NMI), homogeneity score (HS), completeness score (CS), silhouette score (SS), the number of predicted clusters (PC) and the percentage of noise points returned by DBSCAN. Ground truth has 6 clusters. The top-performing scores are shown in boldface.}
\begin{tabular}{llccccccr}
\hline
Model & Clustering Alg. & ARI & NMI & HS & CS & SS & PC & \% Noise \\
\hline
SBERT      & $k$-Means        & 0.9890 & 0.9837 & 0.9837 & 0.9837 & 0.3964 & -- & -- \\
SBERT      & DBSCAN           & 0.5929 & 0.7750 & 0.6921 & 0.8804 & 0.3793 & 5  & 4.72 \\
SBERT      & HDBSCAN + $k$-NN & 0.8185 & 0.9152 & 0.8574 & 0.9813 & 0.4107 & 5  & -- \\
SBERT      & BIRCH            & 0.9909 & 0.9872 & 0.9872 & 0.9872 & 0.3930 & -- & -- \\
\hline
LayoutLMv1 & $k$-Means        & 0.7094 & 0.8530 & 0.8248 & 0.8833 & 0.4763 & -- & -- \\
LayoutLMv1 & DBSCAN           & 0.6910 & 0.8238 & 0.8289 & 0.8187 & 0.4703 & 7  & 7.27 \\
LayoutLMv1 & HDBSCAN + $k$-NN & 0.7327 & 0.8534 & 0.8287 & 0.8796 & 0.4747 & \textbf{6} & -- \\
LayoutLMv1 & BIRCH            & 0.7272 & 0.8662 & 0.8387 & 0.8955 & 0.4767 & -- & -- \\
\hline
LayoutLMv3 & $k$-Means        & 0.7881 & 0.8419 & 0.8415 & 0.8422 & 0.4568 & -- & -- \\
LayoutLMv3 & DBSCAN           & 0.5805 & 0.7205 & 0.7319 & 0.7094 & 0.4522 & 8 & 4.32 \\
LayoutLMv3 & HDBSCAN + $k$-NN & 0.4495 & 0.6455 & 0.8055 & 0.5385 & 0.4131 & 25 & -- \\
LayoutLMv3 & BIRCH            & 0.5597 & 0.6884 & 0.6680 & 0.7101 & 0.5138 & -- & -- \\
\hline
DiT        & $k$-Means        & 0.3795 & 0.5025 & 0.4973 & 0.5079 & 0.3231 & -- & -- \\
DiT        & DBSCAN           & 0.0810 & 0.2647 & 0.1689 & 0.6115 & 0.2278 & 3  & 13.19 \\
DiT        & HDBSCAN + $k$-NN & 0.0483 & 0.2551 & 0.1793 & 0.4415 & 0.1048 & 11 & -- \\
DiT        & BIRCH            & 0.4710 & 0.6091 & 0.5832 & 0.6374 & 0.3260 & -- & -- \\
\hline
Donut      & $k$-Means        & 0.9437 & 0.9471 & 0.9467 & 0.9475 & 0.6253 & -- & -- \\
Donut      & DBSCAN           & 0.8823 & 0.9126 & 0.9702 & 0.8614 & 0.6110 & 9 & 6.55 \\
Donut      & HDBSCAN + $k$-NN & 0.8503 & 0.8790 & 0.9815 & 0.7958 & 0.5352 & 15 & -- \\
Donut      & BIRCH            & 0.9421 & 0.9524 & 0.9514 & 0.9535 & 0.6195 & -- & -- \\
\hline
ColPali    & $k$-Means        & \textbf{1} & \textbf{1} & \textbf{1} & \textbf{1} & 0.5581 & -- & -- \\
ColPali    & DBSCAN           & 0.7337 & 0.8457 & 0.8479 & 0.8435 & 0.5344 & 10 & 8.23 \\
ColPali    & HDBSCAN + $k$-NN & 0.7043 & 0.8583 & 0.7518 & \textbf{1} & 0.6500 & 4  & -- \\
ColPali    & BIRCH            & \textbf{1} & \textbf{1} & \textbf{1} & \textbf{1} & 0.5581 & -- & -- \\
\hline
InternVL3  & $k$-Means        & 0.5195 & 0.6631 & 0.6309 & 0.6987 & \textbf{0.6631} & -- & -- \\
InternVL3  & DBSCAN           & 0.6110 & 0.7172 & 0.7892 & 0.6573 & 0.5934 & 12 & \textbf{2.64} \\
InternVL3  & HDBSCAN + $k$-NN & 0.6364 & 0.8025 & 0.9972 & 0.6714 & 0.5204 & 19 & -- \\
InternVL3  & BIRCH            & 0.5294 & 0.6513 & 0.6365 & 0.6668 & 0.5787 & -- & -- \\
\hline
Gemma3     & $k$-Means        & 0.7853 & 0.8427 & 0.8419 & 0.8435 & 0.6596 & -- & -- \\
Gemma3     & DBSCAN           & 0.6725 & 0.7949 & 0.9354 & 0.6911 & 0.4371 & 17 & 12.15 \\
Gemma3     & HDBSCAN + $k$-NN & 0.7362 & 0.8345 & 0.9945 & 0.7188 & 0.5937 & 17 & -- \\
Gemma3     & BIRCH            & 0.8154 & 0.8725 & 0.8719 & 0.8730 & 0.6556 & -- & -- \\
\bottomrule
\end{tabular}
\label{tab:document_clustering_results}
\end{table}

\paragraph{DBSCAN.} Removing the oracle constraint changes the landscape. Donut now leads the field with ARI = 0.8823, NMI = 0.9126, HS = 0.9702, CS = 0.8614, SS = 0.6110, PC = 9, noise = 6.55\%, i.e., slightly over-segmenting (nine clusters versus six) while maintaining very high homogeneity, a signature of tight density basins that occasionally split a class. ColPali continues to perform well, once more exhibiting slight over-segmentation. Gemma3 achieves ARI = 0.6725, NMI = 0.7949, HS = 0.9354, CS = 0.6911, SS = 0.4371, PC = 17, noise = 0.1215. The sharp divergence between its very high homogeneity and lower completeness indicates over-segmentation into many pure fragments, which aligns with the predicted 17 clusters and non-trivial noise. InternVL3 follows with ARI = 0.6110, NMI = 0.7172, HS = 0.7892, CS = 0.6573, SS = 0.5934, PC = 12, noise = 2.64\%, exhibiting the same over-fragmentation pattern but with less discarded documents. LayoutLMv1 preserves good balance with ARI = 0.6910 and NMI = 0.8238, HS = 0.8289, CS = 0.8187, SS = 0.4703, PC = 7, noise = 0.0727, and LayoutLMv3 is somewhat lower with ARI = 0.5805 and NMI = 0.7205. SBERT drops markedly compared to oracle $k$, achieving ARI = 0.5929, NMI = 0.7750, HS = 0.6921, CS = 0.8804, SS = 0.3793, PC = 5, noise = 4.72\%, showing under-segmentation. DiT degrades most (ARI = 0.0810, NMI = 0.2647), highlighting the brittleness of purely visual tokens when a single global density scale must explain heterogeneous categories.

\paragraph{HDBSCAN + $k$-NN.} The hybrid density-then-assignment scheme eliminates outliers by design and exposes a clear split between under-segmentation and over-segmentation behaviors depending on the embedding. SBERT returns ARI = 0.8185, NMI = 0.9152, HS = 0.8574, CS = 0.9813, SS = 0.4107, PC = 5. The near-perfect completeness with fewer than six clusters indicates merging of classes, consistent with its DBSCAN failure mode. LayoutLMv1 remains balanced with ARI = 0.7327 and NMI = 0.8534, effectively recovering the oracle cardinality. ColPali collapses to four clusters and markedly lower homogeneity, a canonical signature of under-segmentation. In contrast, Donut over-segments (ARI = 0.8503, NMI = 0.8790, HS = 0.9815, CS = 0.7958, SS = 0.5352, PC = 15), creating many very pure clusters that fragment classes. The same pattern is pronounced for the two large vision-language models. Gemma3 yields HS = 0.9945 and CS = 0.7188, and InternVL3 obtains HS = 0.9972 and CS = 0.6714, both showing extreme over-segmentation with near-unity homogeneity but depressed completeness. LayoutLMv3 also over-segments sharply, while DiT performs the worst (ARI = 0.0483, NMI = 0.2551), again reflecting the limitations of visual tokens without textual anchors in this multi-corpus setting.

\begin{figure}[ht]
\centering
\captionsetup[subfigure]{justification=centering}
% ---------------------------------------------------- 1st row
\begin{subfigure}[t]{0.30\textwidth}
  \centering
  \includegraphics[width=\linewidth]{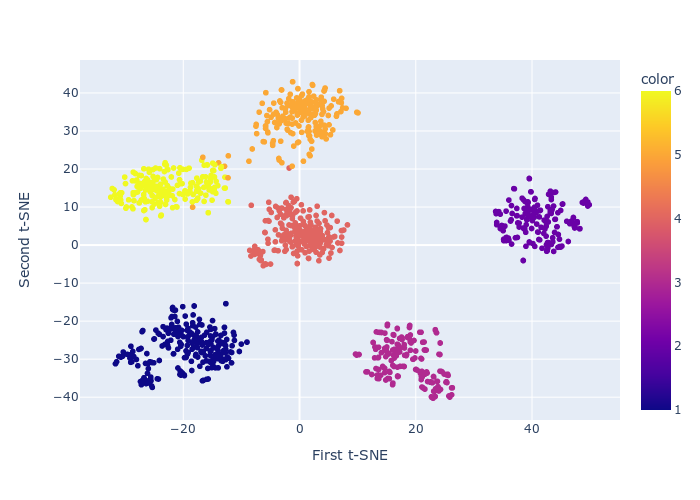}
  \caption{SBERT}
  \label{fig:tsne_sbert}
\end{subfigure}
\hfill
\begin{subfigure}[t]{0.30\textwidth}
  \centering
  \includegraphics[width=\linewidth]{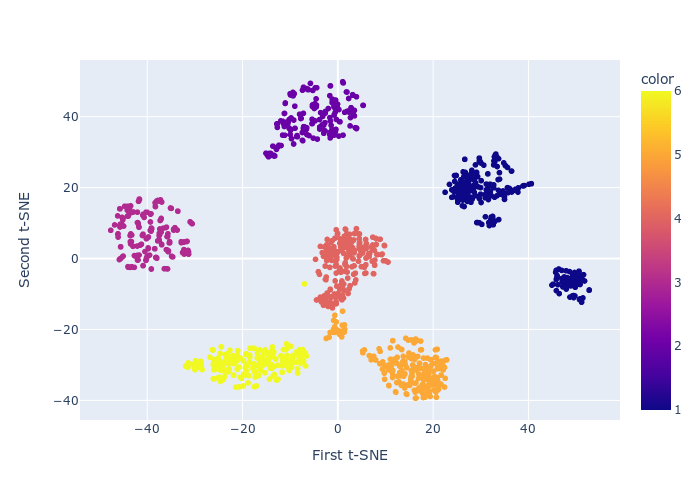}
  \caption{LayoutLMv1}
  \label{fig:tsne_layoutlmv1}
\end{subfigure}
\hfill
\begin{subfigure}[t]{0.30\textwidth}
  \centering
  \includegraphics[width=\linewidth]{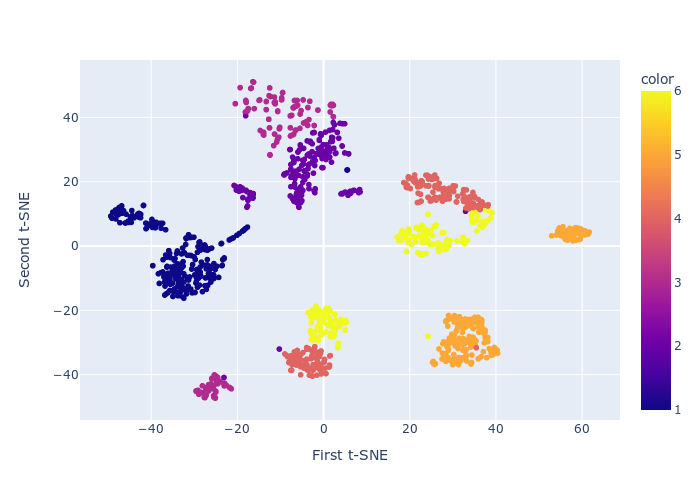}
  \caption{LayoutLMv3}
  \label{fig:tsne_layoutlmv3}
\end{subfigure}

\vspace{0.8em}

% ---------------------------------------------------- 2nd row
\begin{subfigure}[t]{0.30\textwidth}
  \centering
  \includegraphics[width=\linewidth]{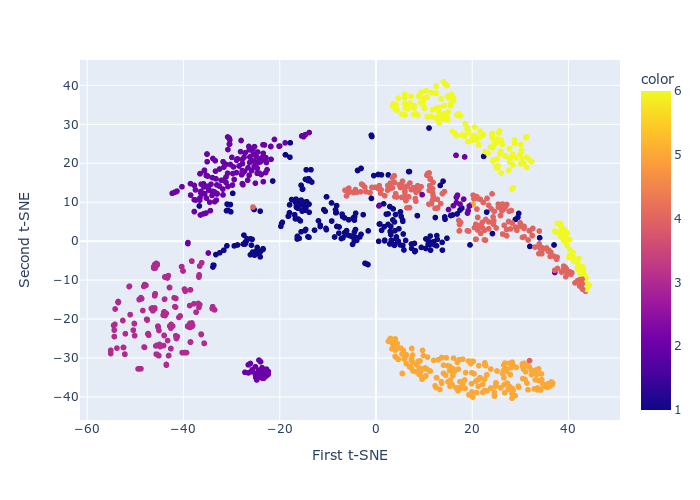}
  \caption{DiT}
  \label{fig:tsne_dit}
\end{subfigure}
\hfill
\begin{subfigure}[t]{0.30\textwidth}
  \centering
  \includegraphics[width=\linewidth]{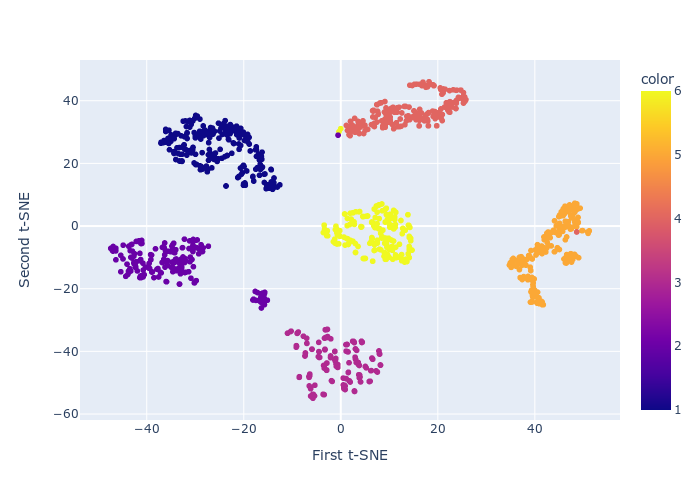}
  \caption{Donut}
  \label{fig:tsne_donut}
\end{subfigure}
\hfill
\begin{subfigure}[t]{0.30\textwidth}
  \centering
  \includegraphics[width=\linewidth]{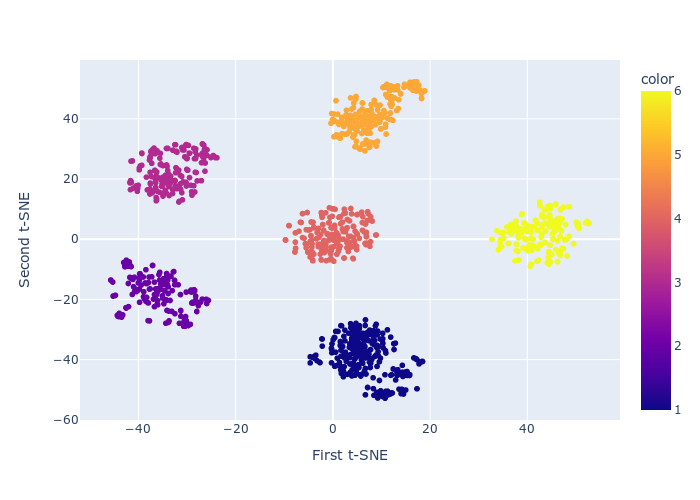}
  \caption{ColPali}
  \label{fig:tsne_colpali}
\end{subfigure}

% ---------------------------------------------------- 3rd row
\begin{subfigure}[t]{0.30\textwidth}
  \centering
  \includegraphics[width=\linewidth]{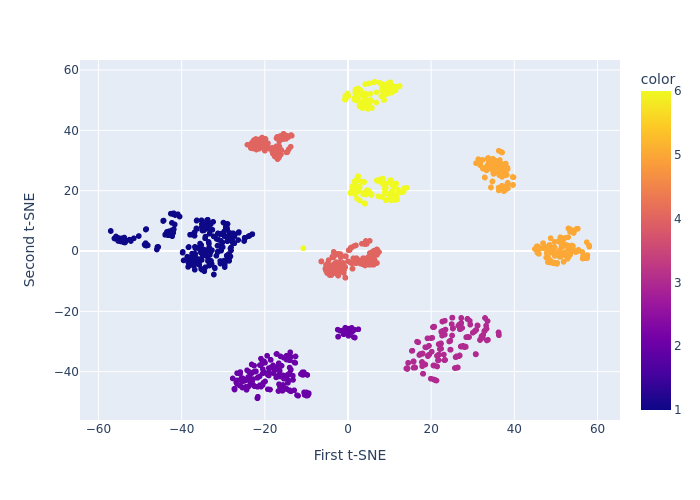}
  \caption{Gemma3}
  \label{fig:tsne_gemma3}
\end{subfigure}
\hspace{1.2em}
\begin{subfigure}[t]{0.30\textwidth}
  \centering
  \includegraphics[width=\linewidth]{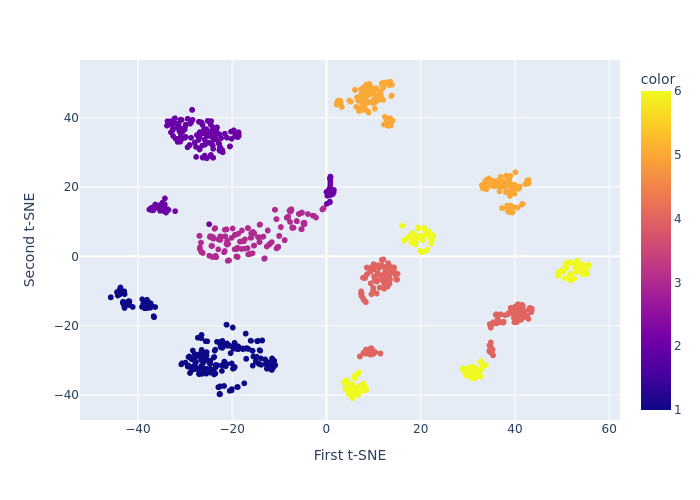}
  \caption{InternVL3}
  \label{fig:tsne_internvl3}
\end{subfigure}

\caption{Two–dimensional t\kern.08em-SNE projections of the mixed-corpus dataset using the eight embedding models. Colors denote document categories; each plot therefore illustrates how well the corresponding representation separates distinct classes in the document-level clustering setting.}
\label{fig:tsne_grid}
\end{figure}

\paragraph{Embedding Geometry via t-SNE.} The qualitative structure revealed by the t-SNE projections in Figure \ref{fig:tsne_grid} mirrors the quantitative rankings reported in Table \ref{tab:document_clustering_results}. Embeddings produced by ColPali and SBERT give rise to compact, approximately isotropic clusters that are cleanly isolated along the manifold, visually corroborating their highest ARI and NMI scores. LayoutLMv1 and Donut retain the same coarse class separation, but their clusters display slight elongation and occasional boundary fraying, consistent with the small performance gap they show relative to the top tier. In the LayoutLMv3 projection, several categories appear fragmented into multiple lobes and radial filaments, anticipating the moderate completeness and homogeneity reported for this model. DiT produces largely amorphous clouds with extensive class overlap, a pattern that echoes its markedly lower document-level clustering scores. 

The projection derived from Gemma3 reveals tightly packed and highly localized clusters, with well-separated class regions and minimal inter-cluster interference. Despite some variation in intra-cluster density, the overall configuration is spatially organized and reflects the model's strong internal cohesion observed in its high silhouette scores, even in settings where it over-segments. In contrast, InternVL3 exhibits more complex geometry: while many clusters remain distinct, the configuration is less uniform, with several class fragments positioned at varying distances and some evidence of discontinuities. This matches its tendency to form compact, low-noise clusters that only partially align with ground-truth labels, as captured by its moderate ARI and NMI values. Together, the t-SNE projections of these newer models confirm that they generate internally coherent structures, though their global label alignment varies according to the clustering objective.

The geometric organization observed in two dimensions therefore reinforces the conclusion that visual-textual fusion and text-centric representations dominate document-level discrimination, whereas the purely visual DiT encoder remains ill-suited to this task.

\paragraph{Interpreting Document-Level Clustering Dynamics.} The document-level corpus exposes three clear patterns. First, when provided with the oracle number of clusters, $k$-Means and BIRCH make the task almost trivial for language‑centric or well‑aligned vision–language embeddings: ColPali is exact (all external metrics = 1), SBERT is nearly perfect, and Donut is close behind.

Second, removing the oracle reveals modality‑dependent failure modes. DBSCAN cleanly separates under‑segmentation from over‑segmentation: SBERT and ColPali tend to merge classes (five and ten clusters, respectively), whereas Donut, Gemma3, and InternVL3 fragment classes (roughly 9, 17, and 12 clusters), often with high homogeneity but reduced completeness. Noise remains moderate for top models (e.g., Donut = 6.6\%, SBERT = 4.7\%) and higher for Gemma3 and DiT (12\% and 13\%, respectively). Even when fragmented, Donut and InternVL3 retain high silhouettes, indicating internally coherent pieces.

Third, HDBSCAN + $k$-NN further clarifies the trade‑off between merging and splitting. ColPali under‑segments, while Donut, Gemma3, and InternVL3 produce many very pure islands. LayoutLMv1 is a notable exception, recovering the true cardinality with balanced homogeneity/completeness. The absence of noise highlights that assigning every point typically trades off purity through over-segmentation or stability through class merging.

From a design perspective, these observations argue that document-level clustering benefits most from embeddings that preserve token identity and language, with vision and layout contributing decisively when typography and structure discriminate classes but not when categories differ primarily by content. DiT’s weaker non‑oracle results reinforce this point.

In practical applications, the gap between oracle and non-oracle settings is critical. Since the true number of clusters is rarely known in advance, strong multimodal embeddings must be paired with mechanisms such as robust $k$-estimation, adaptive density calibration, or post hoc consolidation strategies. Ultimately, language-centric or well-aligned vision and language models offer the most dependable foundation for unsupervised category discovery, but their effectiveness depends on careful alignment between the embedding space and the clustering strategy used.

\subsection{Ablation Studies}
\label{subsec:ablation_studies}

\subsubsection{Size of the Encoder Architecture}
\label{subsubsec:encoder_size}

The first ablation investigates how enlarging LayoutLMv1 from the 113M-parameter base checkpoint to the 340M-parameter large variant affects document-level clustering, where every single-page document is assigned to one of six categories drawn from four distinct datasets. As in the preceding analyses, we contrast oracle k-Means with DBSCAN, the latter operating without knowledge of the true cluster count. Table~\ref{tab:layoutlmv1_ablation_encoder_size} shows the clustering performance obtained with each architecture.

\begin{table}[ht]
\centering
\caption{Ablation study: LayoutLMv1\,\textit{base} versus LayoutLMv1\,\textit{large} on mixed-corpus document-level clustering. Metrics are adjusted rand index (ARI), normalized mutual information (NMI), homogeneity score (HS), completeness score (CS), silhouette score (SS), the number of predicted clusters (PC) and the percentage of noise points returned by DBSCAN. Ground truth has 6 clusters. The top-performing scores are shown in boldface.}
\begin{tabular}{llccccccr}
\hline
Model & Clustering Alg. & ARI & NMI & HS & CS & SS & PC & \% Noise \\
\hline
LayoutLMv1 base  & $k$-Means & 0.7094          & 0.8530 & 0.8248 & 0.8833 & \textbf{0.4763} & -- & -- \\
LayoutLMv1 base  & DBSCAN    & 0.6910          & 0.8238 & 0.8289 & 0.8187 & 0.4703 & \textbf{7} & 7.27 \\
LayoutLMv1 large & $k$-Means & \textbf{0.8692} & \textbf{0.8908} & \textbf{0.8877} & \textbf{0.8908} & 0.4301 & -- & -- \\
LayoutLMv1 large & DBSCAN    & 0.5951          & 0.7845 & 0.7018 & 0.8893 & 0.4701 & \textbf{5} & \textbf{5.12} \\
\bottomrule
\end{tabular}
\label{tab:layoutlmv1_ablation_encoder_size}
\end{table}

Under oracle $k$-Means the larger network provides the expected margin gain: adjusted Rand index rises from 0.71 to 0.87, normalized mutual information increases from 0.85 to 0.89 and both homogeneity and completeness surpass 0.88, confirming that additional depth sharpens the global separation among dataset categories. However, when the number of clusters is unknown, the results are reversed. On DBSCAN every metric except completeness is worse for the large model. The base representation yields 0.69 ARI, 0.82 NMI, 0.83 homogeneity, 0.82 completeness, a silhouette of 0.48, seven clusters discovered, and 7.3\% noise. Scaling up collapses two classes into shared density basins, leaving five clusters, driving ARI down to 0.60, NMI to 0.78, and homogeneity to 0.70, while reducing noise only marginally to 5.1\%.

These results show that, while larger models improve the overall separation between classes, they also smooth out the fine details in the data that density-based methods such as DBSCAN rely on. The divergence between completeness and the other metrics underscores the importance of joint evaluation. Practically, the results caution that scaling a multimodal transformer is beneficial when the cluster cardinality is known, but may hinder open-world discovery unless multi-scale pooling preserves local density structure.

\subsubsection{Mean Pooling Versus \texttt{[CLS]} Token}
\label{subsubsec:mean_pooling_versus_cls}

The second ablation contrasts two ways of compressing the LayoutLMv1 sequence output into a single document vector. The canonical approach retains the first position, the \texttt{[CLS]} token, whereas the alternative averages all token embeddings from the last hidden layer (mean-pool). Both variants were evaluated with oracle $k$-Means and with DBSCAN on the document-level clustering task. Table~\ref{tab:layoutlmv1_ablation_meanpool_versus_token} contrasts the clustering performance obtained with mean-pooled and \texttt{[CLS]}-based LayoutLMv1 embeddings.

\begin{table}[ht]
\centering
\caption{Ablation study: mean pooling versus \texttt{[CLS]} token on mixed-corpus document-level clustering. Metrics are adjusted rand index (ARI), normalized mutual information (NMI), homogeneity score (HS), completeness score (CS), silhouette score (SS), the number of predicted clusters (PC) and the percentage of noise points returned by DBSCAN. Ground truth has 6 clusters. The top-performing scores are shown in boldface.}
\begin{tabular}{llccccccr}
\hline
Approach & Clustering Alg. & ARI & NMI & HS & CS & SS & PC & \% Noise \\
\hline
Mean pooling   & $k$-Means & \textbf{0.7094} & \textbf{0.8530} & 0.8248 & \textbf{0.8833} & \textbf{0.4763} & -- & -- \\
Mean pooling   & DBSCAN    & 0.6910 & 0.8238 & \textbf{0.8289} & 0.8187 & 0.4703 & \textbf{7} & \textbf{7.27} \\
\texttt{[CLS]} & $k$-Means & 0.5513 & 0.6515 & 0.6475 & 0.6556 & 0.5272 & -- & -- \\
\texttt{[CLS]} & DBSCAN    & 0.4658 & 0.6332 & 0.6308 & 0.6357 & 0.4932 & \textbf{7} & 13.35 \\
\hline
\end{tabular}
\label{tab:layoutlmv1_ablation_meanpool_versus_token}
\end{table}

When the number of clusters $k$ is supplied, mean pooling reaches 0.7094 ARI and 0.853 NMI, whereas the \texttt{[CLS]} vector attains only 0.5513 ARI and 0.6515 NMI. Homogeneity and completeness decrease from 0.8248 and 0.8833 to 0.6475 and 0.6556, respectively. With DBSCAN the gap widens: mean pooling preserves 0.691 ARI and 0.8238 NMI, while the \texttt{[CLS]} variant falls to 0.4658 ARI, 0.6332 NMI.

The deficit of the \texttt{[CLS]} approach stems from the fact that LayoutLMv1 is pre-trained with masked-language modelling rather than an explicit document-classification objective. Therefore, the first token learns to encode broad contextual semantics but is not pressured to carve class-conditional margins in embedding space. Mean pooling, by aggregating every token, retains lexical and positional irregularities that remain informative even without supervision.

The ablation shows that without supervised fine-tuning a global \texttt{CLS} summary under-specializes for unsupervised discovery, whereas a simple mean vector offers richer local structure and therefore superior clustering when the number of clusters is unknown. Hybrid representations that concatenate \texttt{[CLS]} with a compressed mean vector may reconcile global semantics with the density variation essential to open-world clustering algorithms.

\subsection{Limitations and Study Validity}
\label{subsec:limitations}

The proposed pipeline achieves strong clustering performance in different types of datasets. Nevertheless, certain assumptions and design choices warrant careful consideration with regard to robustness and operational deployment in broader scenarios. This section outlines the main limitations and, where feasible, delineates concrete and implementable mitigation strategies that remain aligned with the unsupervised orientation of the method.

\subsubsection{Generalization Beyond Single-Page, Well-Structured Layouts}

Our evaluation focuses on single-page, visually well-structured templates such as invoices, identity documents, and certificates. This operating regime is intentionally constrained: the documents exhibit strong layout regularities, a limited variety of page roles, and a high density of stable visual anchors (logos, headers, field grids). Consequently, the learned embedding geometry and the downstream density-based clustering behave favorably. Generalization to longer, loosely structured, or mixed-content documents---such as multi-page reports, letters, contracts, or documents interleaving paragraphs and tables---remains an open question. In these cases, the absence of a rigid field framing and the presence of discourse-level dependencies create two systematic risks: first, that page-level embeddings fail to capture document-level semantics; second, that token and patch counts induce scale effects that distort densities in the representation space, thereby confounding density-based clustering thresholds.

To mitigate these issues while preserving the unsupervised character of the method, a plausible mitigation is a hierarchical aggregation over a per-document graph that integrates pagination and semantic affinity before producing a single document-level representation. The construction proceeds as follows. For a document comprising \(n\) pages, we first compute page embeddings \(h_i \in \mathbb{R}^d\) using the same encoders as in the single-page pipeline (textual, visual, or multimodal). We then instantiate an undirected weighted graph \(G=(V,E)\) with one node per page. Edges encode two complementary notions of relatedness. Sequential edges connect consecutive pages and enforce pagination; they are assigned a constant structural weight \(\lambda_{\mathrm{seq}}>0\). Semantic edges connect non-adjacent pages that are similar in content; for each page \(i\), we select its top-\(k\) most similar non-adjacent pages by cosine similarity in embedding space and connect them with weight \(\lambda_{\mathrm{sim}}\max(0,\cos(h_i,h_j))\), with \(\lambda_{\mathrm{sim}}>0\). To ensure symmetry without discarding useful links, we use a symmetric \(k\)-nearest-neighbor rule over non-adjacent pairs: an edge \((i,j)\) is present if either \(i\) selects \(j\) or \(j\) selects \(i\). Self-loops are set to unit weight to stabilize subsequent propagation. Collecting the edge weights \(w_{ij}\) into the weighted adjacency \(A=[w_{ij}]\) and degrees \(d_i=\sum_j w_{ij}\) into \(D=\mathrm{diag}(d_1,\dots,d_n)\), we form the symmetric normalized adjacency
\[
\tilde{A}=D^{-\frac{1}{2}} A D^{-\frac{1}{2}}.
\]
Normalization is necessary to balance the influence of nodes with different degrees, to obtain a spectrally bounded propagation operator, and to render the resulting document representation comparable across documents of different lengths. One or two steps of Laplacian smoothing are then applied to denoise page embeddings and inject pagination and long-range semantic context,
\[
\hat{H}=\tilde{A}H \quad \text{or} \quad \hat{H}=\tilde{A}(\tilde{A}H),
\]
where \(H=[h_1,\dots,h_n]^\top\). We next compute a provisional document prototype \(\bar{h} = \frac{1}{n}\sum_{i=1}^n \hat{h}_i\) and derive unsupervised salience weights by a temperature-scaled cosine attention,
\[
\alpha_i=\frac{\exp(\cos(\hat{h}_i,\bar{h})/\tau)}{\sum_{j=1}^n \exp(\cos(\hat{h}_j,\bar{h})/\tau)}, \qquad \tau>0.
\]
The final document embedding is obtained by length-normalized, salience-weighted pooling,
\[
h_{\mathrm{doc}}=\sum_{i=1}^n \alpha_i \hat{h}_i,
\]
followed by \(\ell_2\)-normalization prior to clustering. This hierarchical procedure preserves local page evidence, integrates inter-page dependencies through graph propagation, and counteracts scale effects via degree normalization and attention-based weighting.

A five-page example illustrates the construction and the role of normalization. Consider pages indexed \(1\) to \(5\). We set \(\lambda_{\mathrm{seq}}=1\), \(\lambda_{\mathrm{sim}}=0.5\), and \(k=1\) for the semantic non-adjacent links. Adjacent pairs \((1,2)\), \((2,3)\), \((3,4)\), and \((4,5)\) receive the structural weight \(1\), and all pages have self-loops of weight \(1\). Suppose that, among non-adjacent pairs, page \(1\) is most similar to page \(3\) with \(\cos(h_1,h_3)=0.8\), page \(2\) is most similar to page \(4\) with \(\cos(h_2,h_4)=0.6\), and page \(5\) is most similar to page \(3\) with \(\cos(h_5,h_3)=0.7\). The symmetric \(k\)-nearest-neighbor rule therefore adds non-adjacent semantic edges \((1,3)\), \((2,4)\), and \((3,5)\) with weights \(0.4\), \(0.3\), and \(0.35\), respectively. The resulting weighted adjacency is given by
\[
A=\begin{bmatrix}
1 & 1 & 0.4 & 0 & 0 \\
1 & 1 & 1 & 0.3 & 0 \\
0.4 & 1 & 1 & 1 & 0.35 \\
0 & 0.3 & 1 & 1 & 1 \\
0 & 0 & 0.35 & 1 & 1
\end{bmatrix}.
\]
The degree matrix is \(D=\mathrm{diag}(2.4,\,3.3,\,3.75,\,3.3,\,2.35)\). The symmetric normalized adjacency \(\tilde{A}=D^{-1/2}AD^{-1/2}\) is then
\[
\tilde{A}\approx
\begin{bmatrix}
0.416667 & 0.355335 & 0.133333 & 0         & 0 \\
0.355335 & 0.303030 & 0.284268 & 0.090909  & 0 \\
0.133333 & 0.284268 & 0.266667 & 0.284268  & 0.117901 \\
0        & 0.090909 & 0.284268 & 0.303030  & 0.359095 \\
0        & 0        & 0.117901 & 0.359095  & 0.425532
\end{bmatrix}.
\]
Applying \(\hat{H}=\tilde{A}H\) propagates information along both the pagination chain and the added non-adjacent semantic links while ensuring that pages with higher raw degree do not dominate the update merely due to connectivity. The attention weights \(\alpha_i\) subsequently emphasize pages whose smoothed embeddings are most representative of the document prototype, thus yielding a single \(h_{\mathrm{doc}}\) that is invariant to length and robust to heterogeneous page informativeness. The entire procedure is unsupervised, computationally light—since operations are linear in the number of pages—and integrates seamlessly with the downstream clustering stage.

\subsubsection{Operating under Weak and Semi-Supervision}

The proposed method is purposefully fully unsupervised. Nevertheless, industrial IDP pipelines often operate under limited supervision, such as sparse labels, heuristic keyword rules, or implicit metadata. A principled compromise is to remain unsupervised in representation learning and initial clustering while admitting strictly weak signals at consolidation time. One practical strategy is \emph{constraint-aware consolidation} in which soft \emph{must-link} and \emph{cannot-link} hints are derived automatically from inexpensive sources; for example, pages sharing a stable issuer string or a vendor identifier induce soft must-links, whereas mutually exclusive headers or incompatible jurisdiction codes yield soft cannot-links. These constraints do not alter the initial clusters, rather they guide a post hoc reconciliation step that resolves conflicts by minimizing a small penalty over cluster assignments subject to the discovered partition. 

An adjacent strategy is \emph{prototype seeding with budgeted exemplars}. A small, fixed budget B of highly representative documents (medoids) is selected---typically one per putative class---to serve as \emph{seed prototypes}. These seeds are held fixed, and a single seeded refinement is performed: every non-seed point is reassigned to the nearest seed prototype (with a mild inertia term to discourage large jumps and with any cannot-links enforced), after which prototypes are optionally updated once by averaging assigned points. This one-pass anchoring collapses fragmented micro-clusters around a handful of trusted exemplars, requires no full labels, preserves the unsupervised character of the method, and adds only linear-time overhead.

\subsubsection{Post-Clustering Consolidation and Adaptive Estimation of the Number of Clusters}
Empirically, some encoders occasionally produce over-segmented partitions; large vision--language models may yield many fine-grained, internally pure clusters that exceed the number of ground-truth classes. The current pipeline does not include a consolidation mechanism or an adaptive procedure to estimate the effective number of clusters and therefore may present inflated fragmentation that complicates downstream consumption.

Two complementary refinements are practical and computationally light. First, \emph{centroid-space agglomeration} provides a deterministic and fast consolidation layer. After base clustering, one operates in the space of cluster centroids and fits a simple mixture model (for example, a Gaussian mixture with shared spherical covariance), selecting the number of merged components by an information criterion such as Bayesian Information Criterion (BIC). The resulting centroid assignments induce a merge map that is applied to the original clusters. Because the operation is performed on centroids, the computational cost is negligible relative to the base clustering, and the risk of over-merging is controlled by an explicit model-selection criterion rather than a heuristic threshold.

Second, \emph{stability-guided merging} formalizes consolidation as a test of persistence under sampling variability. One repeatedly subsamples documents, reclusters with fixed hyperparameters, and computes a co-association matrix. Clusters whose centroids are close and whose members consistently co-occur across bootstraps are merged, whereas unstable micro-clusters are pruned or reassigned to the nearest stable macro-cluster based on a local density ratio. This procedure is well suited to density-based methods and yields a partition that is robust to dataset perturbations.

Adaptive estimation of the number of clusters is attainable without ground truth by combining the two ideas above. One first reduces over-segmentation via stability-guided merging and then selects the final \(k\) by maximizing an internal validity index (such as silhouette) on the merged partition, evaluated in centroid space to avoid bias from variable cluster sizes. In practice, this produces an explicit estimate of \(k\) together with a quantification of uncertainty through the dispersion of bootstrap-derived \(k\) values, thereby replacing oracle \(k\) with a defensible, data-driven quantity.

\subsubsection{Robustness to OCR Errors, Multilingual Data, and Real-World Variability}
The current approach leverages textual signals obtained by OCR alongside layout and vision cues. In operational settings, OCR quality varies with scan resolution, compression, and handwriting; furthermore, documents may be multilingual or contain mixed scripts. These factors perturb textual embeddings, shift the joint geometry of text--layout--vision representations, and degrade the stability of density-based clustering algorithms.

A practical route to robustness is \emph{multi-view fusion with confidence-aware weighting}. Independent text-only and vision-only embeddings are computed and combined via late fusion in which the contribution of the text view is modulated by OCR confidence statistics (for example, mean character confidence, or proportion of low-confidence tokens). When OCR is unreliable, the fusion weights tilt toward the vision channel. When the image is degraded, but the OCR remains reliable, the fusion prioritizes the text channel. This strategy does not require retraining encoders and can be implemented as a convex combination in embedding space followed by the existing clustering stage.

Multilingual inputs and mixed scripts often introduce language-specific shifts in embedding means and covariances—exacerbated by OCR variability that cause clusters to separate by language instead of template. When such behavior is observed, first apply a lightweight language and script identification module to the OCR text. Retain a single multilingual encoder and normalize the embeddings separately by language. For each language, estimate batch-level mean and covariance, and perform either whitening or CORAL alignment \cite{Sun2016CORAL} to a chosen reference domain. Finally, perform clustering on these aligned representations, optionally fusing text and vision/layout embeddings with weights modulated by OCR confidence so that poor transcripts do not dominate. This training-free and unsupervised normalization reduces cross-language geometry drift while preserving template-level semantics. It should be applied only when language-driven fragmentation is empirically detected and omitted when multilingual embeddings are already well aligned or when language itself is the clustering objective.

\section{Conclusions}
\label{sec:conclusions}

This work presented an unsupervised pipeline for clustering documents at both the category and template levels by combining frozen multimodal encoders with classical clustering algorithms. Specifically, we evaluated four distinct clustering methods: $k$-Means, DBSCAN, a combination of HDBSCAN with $k$-NN, and BIRCH. Together, these algorithms allowed us to probe both centroid-based and density-based grouping behaviors under oracle and non-oracle settings. Experiments were conducted on five diverse datasets, including synthetic invoices and noisy scanned pages, and revealed consistent patterns in model behavior.

Visual encoders such as DiT and the vision component of Donut almost perfectly separate templates when images are clean and well aligned, achieving adjusted-Rand scores well above 0.95. However, their performance degrades sharply when illumination, resolution, or geometry are heavily perturbed. In contrast, text-centric SBERT and multimodal retrieval-oriented ColPali remain remarkably stable under the same visual stress, preserving cluster integrity even when the page is heavily distorted, although they trail vision models on pristine data. Hybrid encoders that are pre-trained with different modalities (tokens, spatial coordinates, and pixels), such as LayoutLMv3, Donut, Gemma3, and InternVL3, consistently offer the best trade-off: they approach the vision models on clean images while retaining most of the robustness of the language models under moderate noise. Among them, Donut yields the most compact manifolds and the highest fully unsupervised accuracy on document-level tasks, despite the absence of an explicit layout-alignment loss.

Ablation studies revealed that simple modality-aware mean pooling outperforms conventional \texttt{[CLS]} token approach and that scaling encoder size yields diminishing returns. From these results, five key conclusions emerge. First, template clustering benefits from using all three modalities — text, layout, and vision — since each dominates in different regions of the document space. Second, document category clustering is also improved through multimodal fusion, where token identity offers a strong starting point, and layout and visual cues help resolve harder cases. Third, Donut, Gemma3 and InternVL3 serve as effective alternatives to established models, producing compact clusters even without layout-specific supervision. Fourth, classical clustering algorithms can perform competitively when combined with strong pre-trained embeddings, reducing the need for task-specific fine-tuning. Fifth, density-based algorithms uncover modality-specific clustering errors, such as merging in text models or over-fragmentation in vision-first models, which are not visible when the number of clusters is assumed in advance.

\bibliographystyle{unsrt}
\bibliography{references}

\begin{thebibliography}{10}

\bibitem{Cozzolino2022}
Ilaria Cozzolino and Marco~Bertini Ferraro.
\newblock Document clustering.
\newblock {\em WIREs Computational Statistics}, 14(6):e1588, nov 2022.

\bibitem{Yan2017}
Wei Yan, Bob Zhang, Sihan Ma, and Zuyuan Yang.
\newblock A novel regularized concept factorization for document clustering.
\newblock {\em Knowledge-Based Systems}, 135:147--158, nov 2017.

\bibitem{Diallo2021}
Bassoma Diallo, Jie Hu, Tianrui Li, Ghufran~Ahmad Khan, Xinyan Liang, and Yimiao Zhao.
\newblock Deep embedding clustering based on contractive autoencoder.
\newblock {\em Neurocomputing}, 433:96--107, apr 2021.

\bibitem{Devlin2019}
Jacob Devlin, Ming-Wei Chang, Kenton Lee, and Kristina Toutanova.
\newblock {BERT}: Pre-training of deep bidirectional transformers for language understanding.
\newblock In {\em Proceedings of the 2019 Conference of the North American Chapter of the Association for Computational Linguistics: Human Language Technologies, Volume 1 (Long and Short Papers)}, pages 4171--4186. Association for Computational Linguistics, 2019.

\bibitem{Liu2019}
Yinhan Liu, Myle Ott, Naman Goyal, Jingfei Du, Mandar Joshi, Danqi Chen, Omer Levy, Mike Lewis, Luke Zettlemoyer, and Veselin Stoyanov.
\newblock {RoBERTa}: A robustly optimized {BERT} pretraining approach.
\newblock {\em arXiv:1907.11692}, 2019.

\bibitem{PETUKHOVA2025100}
Alina Petukhova, João~P. Matos-Carvalho, and Nuno Fachada.
\newblock Text clustering with large language model embeddings.
\newblock {\em International Journal of Cognitive Computing in Engineering}, 6:100--108, 2025.

\bibitem{kLLMmeans2025}
Jairo Diaz-Rodriguez.
\newblock {k-LLMmeans:} scalable, stable, and interpretable text clustering via {LLM}-based centroids.
\newblock {\em arXiv:2502.09667}, 2025.

\bibitem{forgy1965}
Edward~W. Forgy.
\newblock Cluster analysis of multivariate data: efficiency versus interpretability of classifications.
\newblock {\em Biometrics}, 21:768--780, 1965.

\bibitem{lloyd1982}
Stuart~P. Lloyd.
\newblock Least squares quantization in {PCM}.
\newblock {\em IEEE Transactions on Information Theory}, 28(2):129--137, 1982.

\bibitem{ahmed2020}
Mohiuddin Ahmed, Raihan Seraj, and Syed Mohammed~Shamsul Islam.
\newblock The k-means algorithm: A comprehensive survey and performance evaluation.
\newblock {\em Electronics}, 9(8), 2020.

\bibitem{ester1996}
Martin Ester, Hans-Peter Kriegel, J\"{o}rg Sander, and Xiaowei Xu.
\newblock A density-based algorithm for discovering clusters in large spatial databases with noise.
\newblock In {\em Proceedings of the Second International Conference on Knowledge Discovery and Data Mining}, KDD'96, pages 226--231. AAAI Press, 1996.

\bibitem{campello2013}
Ricardo J. G.~B. Campello, Davoud Moulavi, and Joerg Sander.
\newblock Density-based clustering based on hierarchical density estimates.
\newblock In Jian Pei, Vincent~S. Tseng, Longbing Cao, Hiroshi Motoda, and Guandong Xu, editors, {\em Advances in Knowledge Discovery and Data Mining}, pages 160--172, Berlin, Heidelberg, 2013. Springer Berlin Heidelberg.

\bibitem{FixHodges1951}
Evelyn Fix and Joseph~L. Hodges.
\newblock Discriminatory analysis. nonparametric discrimination: consistency properties.
\newblock {\em Technical Report 4. Project 21-49-004, USAF School of Aviation Medicine, Randolph Field, Texas}, 1951.
\newblock Reprinted in International Statistical Review, 57(3), 1989, pp. 238–247.

\bibitem{birch1996}
Tian Zhang, Raghu Ramakrishnan, and Miron Livny.
\newblock {BIRCH}: An efficient data clustering method for very large databases.
\newblock In {\em Proceedings of the 1996 ACM SIGMOD International Conference on Management of Data}, SIGMOD '96, pages 103--114, New York, NY, USA, 1996. Association for Computing Machinery.

\bibitem{reimers2019sbert}
Nils Reimers and Iryna Gurevych.
\newblock {Sentence-BERT}: Sentence embeddings using siamese {BERT}-networks.
\newblock In {\em Proceedings of the 2019 Conference on Empirical Methods in Natural Language Processing (EMNLP) and the 9th International Joint Conference on Natural Language Processing (IJCNLP)}, pages 3982--3992. Association for Computational Linguistics, 2019.

\bibitem{Xu2020}
Yiheng Xu, Minghao Li, Lei Cui, Shaohan Huang, Furu Wei, and Ming Zhou.
\newblock Layout{LM}: Pre-training of text and layout for document image understanding.
\newblock In {\em Proceedings of the 26th ACM SIGKDD International Conference on Knowledge Discovery \& Data Mining}, KDD '20, pages 1192–--1200, New York, NY, USA, 2020. Association for Computing Machinery.

\bibitem{Xu2022}
Yupan Huang, Tengchao Lv, Lei Cui, Yutong Lu, and Furu Wei.
\newblock Layout{LM}v3: Pre-training for document {AI} with unified text and image masking.
\newblock In {\em Proceedings of the 30th ACM International Conference on Multimedia}, MM '22, pages 4083--4091, New York, NY, USA, 2022. Association for Computing Machinery.

\bibitem{Li2022}
Junlong Li, Yiheng Xu, Tengchao Lv, Lei Cui, Cha Zhang, and Furu Wei.
\newblock {DiT}: Self-supervised pre-training for document image transformer.
\newblock In {\em Proceedings of the 30th ACM International Conference on Multimedia}, MM '22, pages 3530--3539, New York, NY, USA, 2022. Association for Computing Machinery.

\bibitem{Kim2022}
Geewook Kim, Teakgyu Hong, Moonbin Yim, JeongYeon Nam, Jinyoung Park, Jinyeong Yim, Wonseok Hwang, Sangdoo Yun, Dongyoon Han, and Seunghyun Park.
\newblock {OCR}-free document understanding transformer.
\newblock In Shai Avidan, Gabriel Brostow, Moustapha Ciss{\'e}, Giovanni~Maria Farinella, and Tal Hassner, editors, {\em Computer Vision -- ECCV 2022}, pages 498--517, Cham, 2022. Springer Nature Switzerland.

\bibitem{Faysse2025}
Manuel Faysse, Hugues Sibille, Tony Wu, Bilel Omrani, Gautier Viaud, Céline Hudelot, and Pierre Colombo.
\newblock {ColPali}: Efficient document retrieval with vision language models.
\newblock {\em arXiv:2407.01449}, 2024.

\bibitem{gemma32025}
Gemma Team, Aishwarya Kamath, Johan Ferret, Shreya Pathak, Nino Vieillard, Ramona Merhej, et~al.
\newblock Gemma 3 technical report.
\newblock {\em arXiv:2503.19786}, 2025.

\bibitem{internvl32025}
Jinguo Zhu, Weiyun Wang, Zhe Chen, Zhaoyang Liu, Shenglong Ye, Lixin Gu, et~al.
\newblock {InternVL3}: Exploring advanced training and test-time recipes for open-source multimodal mmodels.
\newblock {\em arXiv:2504.10479}, 2025.

\bibitem{Xie2016}
Junyuan Xie, Ross Girshick, and Ali Farhadi.
\newblock Unsupervised deep embedding for clustering analysis.
\newblock In Maria~Florina Balcan and Kilian~Q. Weinberger, editors, {\em Proceedings of The 33rd International Conference on Machine Learning}, volume~48 of {\em Proceedings of Machine Learning Research}, pages 478--487, New York, New York, USA, 20--22 Jun 2016. PMLR.

\bibitem{RepresentationLearningShortTextClustering}
Hui Yin, Xiangyu Song, Shuiqiao Yang, Guangyan Huang, and Jianxin Li.
\newblock Representation learning for short text clustering.
\newblock In Wenjie Zhang, Lei Zou, Zakaria Maamar, and Lu~Chen, editors, {\em Web Information Systems Engineering -- WISE 2021}, pages 321--335, Cham, 2021. Springer International Publishing.

\bibitem{TextClusteringWeightedBERT}
Yutong Li, Juanjuan Cai, and Jingling Wang.
\newblock A text document clustering method based on weighted {BERT} model.
\newblock In {\em 2020 IEEE 4th Information Technology, Networking, Electronic and Automation Control Conference (ITNEC)}, volume~1, pages 1426--1430, 2020.

\bibitem{TowardsMultimodalMultitask2022}
Subhojeet Pramanik, Shashank Mujumdar, and Hima Patel.
\newblock Towards a multi-modal, multi-task learning based pre-training framework for document representation learning.
\newblock {\em arXiv:2009.14457}, 2022.

\bibitem{Fang2023}
Uno Fang, Man Li, Jianxin Li, Longxiang Gao, Tao Jia, and Yanchun Zhang.
\newblock A comprehensive survey on multi-view clustering.
\newblock {\em IEEE Transactions on Knowledge and Data Engineering}, 35(12):12350--12368, 2023.

\bibitem{Bai2024}
Ruina Bai and Qi~Bai.
\newblock Improving multi-view document clustering: Leveraging multi-structure processor and hybrid ensemble clustering module.
\newblock In {\em Proceedings of the 2024 Joint International Conference on Computational Linguistics, Language Resources and Evaluation (LREC-COLING 2024)}, pages 8866--8876. ELRA and ICCL, 2024.

\bibitem{Zhu2024}
Pengfei Zhu, Xinjie Yao, Yu~Wang, Binyuan Hui, Dawei Du, and Qinghua Hu.
\newblock Multiview deep subspace clustering networks.
\newblock {\em IEEE Transactions on Cybernetics}, 54(7):4280--4293, 2024.

\bibitem{Brbic2018}
Maria Brbić and Ivica Kopriva.
\newblock Multi-view low-rank sparse subspace clustering.
\newblock {\em Pattern Recognition}, 73:247--258, 2018.

\bibitem{Hu2023}
Shizhe Hu, Guoliang Zou, Chaoyang Zhang, Zhengzheng Lou, Ruilin Geng, and Yangdong Ye.
\newblock Joint contrastive triple-learning for deep multi-view clustering.
\newblock {\em Information Processing \& Management}, 60(3):103284, 2023.

\bibitem{Bai2021}
Ruina Bai, Ruizhang Huang, Yanping Chen, and Yongbin Qin.
\newblock Deep multi-view document clustering with enhanced semantic embedding.
\newblock {\em Information Sciences}, 564:273--287, 2021.

\bibitem{Bai2024b}
Ruina Bai, Ruizhang Huang, Yanping Chen, Yongbin Qin, Yong Xu, and Qinghua Zheng.
\newblock A hierarchical consensus learning model for deep multi-view document clustering.
\newblock {\em Information Fusion}, 111:102507, 2024.

\bibitem{Zhao2023}
Xiaojia Zhao, Tingting Xu, Qiangqiang Shen, Youfa Liu, Yongyong Chen, and Jingyong Su.
\newblock Double high-order correlation preserved robust multi-view ensemble clustering.
\newblock {\em ACM Transactions on Multimedia Computing, Communications and Applications}, 20(1):1--21, 2023.

\bibitem{Yin2020}
Ming Yin, Weitian Huang, and Junbin Gao.
\newblock Shared generative latent representation learning for multi-view clustering.
\newblock In {\em Proceedings of the AAAI conference on artificial intelligence}, volume~34, pages 6688--6695, 2020.

\bibitem{Xu2021}
Yang Xu, Yiheng Xu, Tengchao Lv, Lei Cui, Furu Wei, Guoxin Wang, Yijuan Lu, Dinei Florencio, Cha Zhang, Wanxiang Che, Min Zhang, and Lidong Zhou.
\newblock {L}ayout{LM}v2: Multi-modal pre-training for visually-rich document understanding.
\newblock In Chengqing Zong, Fei Xia, Wenjie Li, and Roberto Navigli, editors, {\em Proceedings of the 59th Annual Meeting of the Association for Computational Linguistics and the 11th International Joint Conference on Natural Language Processing (Volume 1: Long Papers)}, pages 2579--2591, Online, August 2021. Association for Computational Linguistics.

\bibitem{Powalski2021}
Rafa{\l} Powalski, {\L}ukasz Borchmann, Dawid Jurkiewicz, Tomasz Dwojak, Micha{\l} Pietruszka, and Gabriela Pa{\l}ka.
\newblock Going full-{TILT} boogie on document understanding with text-image-layout transformer.
\newblock In Josep Llad{\'o}s, Daniel Lopresti, and Seiichi Uchida, editors, {\em {Document Analysis and Recognition -- ICDAR 2021}}, pages 732--747, Cham, 2021. Springer International Publishing.

\bibitem{Li2023}
Minghao Li, Tengchao Lv, Jingye Chen, Lei Cui, Yijuan Lu, Dinei Florencio, Cha Zhang, Zhoujun Li, and Furu Wei.
\newblock {TrOCR}: Transformer-based optical character recognition with pre-trained models.
\newblock In {\em Proceedings of the Thirty-Seventh AAAI Conference on Artificial Intelligence and Thirty-Fifth Conference on Innovative Applications of Artificial Intelligence and Thirteenth Symposium on Educational Advances in Artificial Intelligence}, AAAI'23/IAAI'23/EAAI'23. AAAI Press, 2023.

\bibitem{Appalaraju2021}
Srikar Appalaraju, Bhavan Jasani, Bhargava~Urala Kota, Yusheng Xie, and R.~Manmatha.
\newblock {DocFormer}: End-to-end transformer for document understanding.
\newblock In {\em 2021 IEEE/CVF International Conference on Computer Vision (ICCV)}, pages 973--983, 2021.

\bibitem{Adhikari2019}
Ashutosh Adhikari, Achyudh Ram, Raphael Tang, and Jimmy Lin.
\newblock {DocBERT}: {BERT} for document classification.
\newblock {\em arXiv:1904.08398}, 2019.

\bibitem{LiGT2025}
Thanh-Phong Le, Trung Le~Chi Phan, Nghia~Hieu Nguyen, and Kiet~Van Nguyen.
\newblock {LiGT}: Layout-infused generative transformer for visual question answering on vietnamese receipts.
\newblock {\em arXiv:2502.19202}, 2025.

\bibitem{Lv2023}
Tengchao Lv, Yupan Huang, Jingye Chen, Ming Ding, Zhenfang Xiao, Wen Wang, Zhe Zhao, Li~Dong, and Furu Wei.
\newblock Kosmos-2.5: A multimodal literate model.
\newblock {\em arXiv:2309.11419}, 2023.

\bibitem{Tang2023}
Zineng Tang, Ziyi Yang, Guoxin Wang, Yuwei Fang, Yang Liu, Chenguang Zhu, Michael Zeng, Cha Zhang, and Mohit Bansal.
\newblock Unifying vision, text, and layout for universal document processing.
\newblock In {\em 2023 IEEE/CVF Conference on Computer Vision and Pattern Recognition (CVPR)}, pages 19254--19264, 2023.

\bibitem{Lu2022}
Jiasen Lu, Christopher Clark, Rowan Zellers, Seungwon Lee, Niket Tandon, Jordi Pont-Tuset, and Yejin Choi.
\newblock {Unified-IO}: A unified model for vision, language, and multi-modal tasks.
\newblock {\em arXiv:2206.08916}, 2022.

\bibitem{Liu2024}
Chaohu Liu, Kun Yin, Haoyu Cao, Xinghua Jiang, Xin Li, Yinsong Liu, Deqiang Jiang, Xing Sun, and Linli Xu.
\newblock {HRVDA}: High-resolution visual document assistant.
\newblock In {\em 2024 IEEE/CVF Conference on Computer Vision and Pattern Recognition (CVPR)}, pages 15534--15545, Los Alamitos, CA, USA, 2024. IEEE Computer Society.

\bibitem{pearson1901pca}
Karl Pearson.
\newblock On lines and planes of closest fit to systems of points in space.
\newblock {\em The London, Edinburgh, and Dublin Philosophical Magazine and Journal of Science}, 2(11):559--572, 1901.

\bibitem{siglip2023}
Xiaohua Zhai, Basil Mustafa, Alexander Kolesnikov, and Lucas Beyer.
\newblock Sigmoid loss for language image pre-training.
\newblock In {\em 2023 IEEE/CVF International Conference on Computer Vision (ICCV)}, pages 11941--11952, 2023.

\bibitem{qwen2025}
Qwen, An~Yang, Baosong Yang, Beichen Zhang, Binyuan Hui, Bo~Zheng, et~al.
\newblock {Qwen2.5} technical report.
\newblock {\em arXiv:2412.15115}, 2024.

\bibitem{Shi2017}
Baoguang Shi, Xiang Bai, and Cong Yao.
\newblock An end-to-end trainable neural network for image-based sequence recognition and its application to scene text recognition.
\newblock {\em IEEE Transactions on Pattern Analysis \& Machine Intelligence}, 39(11):2298--2304, November 2017.

\bibitem{pedregosa2011scikit}
Fabian Pedregosa, Gaël Varoquaux, Alexandre Gramfort, Vincent Michel, Bertrand Thirion, Olivier Grisel, Mathieu Blondel, Peter Prettenhofer, Ron Weiss, Vincent Dubourg, Jake Vanderplas, Alexandre Passos, David Cournapeau, Matthieu Brucher, Matthieu Perrot, and Édouard Duchesnay.
\newblock Scikit-learn: machine learning in {Python}.
\newblock {\em Journal of Machine Learning Research}, 12(85):2825--2830, 2011.

\bibitem{steinley2004properties}
Douglas Steinley.
\newblock Properties of the {Hubert-Arabie} adjusted {Rand} index.
\newblock {\em Psychological Methods}, 9(3):386--396, 2004.

\bibitem{vinh2009information}
Nguyen~Xuan Vinh, Julien Epps, and James Bailey.
\newblock Information theoretic measures for clusterings comparison: {Is} a correction for chance necessary?
\newblock In {\em Proceedings of the 26th International Conference on Machine Learning (ICML)}, pages 1073--1080, 2009.

\bibitem{rosenberg2007v}
Andrew Rosenberg and Julia Hirschberg.
\newblock V-measure: A conditional entropy-based external cluster evaluation measure.
\newblock In {\em Proceedings of the 2007 Joint Conference on Empirical Methods in Natural Language Processing and Computational Natural Language Learning (EMNLP-CoNLL)}, pages 410--420, 2007.

\bibitem{rousseeuw1987silhouettes}
Peter~J. Rousseeuw.
\newblock Silhouettes: A graphical aid to the interpretation and validation of cluster analysis.
\newblock {\em Journal of Computational and Applied Mathematics}, 20:53--65, 1987.

\bibitem{limam2025fatura}
Mahmoud Limam, Marwa Dhiaf, and Yousri Kessentini.
\newblock Information extraction from multi-layout invoice images using {FATURA} dataset.
\newblock {\em Engineering Applications of Artificial Intelligence}, 149:110478, 2025.

\bibitem{huangicdar2019}
Zheng Huang, Kai Chen, Jianhua He, Xiang Bai, Dimosthenis Karatzas, Shijian Lu, and C.~V. Jawahar.
\newblock {ICDAR 2019} competition on scanned receipt {OCR} and information extraction.
\newblock In {\em {2019 International Conference on Document Analysis and Recognition (ICDAR)}}, pages 1516--1520, 2019.

\bibitem{soares2020bid}
Álysson Soares, Ricardo das Neves~Junior, and Byron Bezerra.
\newblock {BID Dataset}: {A} challenge dataset for document processing tasks.
\newblock In {\em {Anais Estendidos da XXXIII Conference on Graphics, Patterns and Images}}, pages 143--146, Porto Alegre, RS, Brasil, 2020. SBC.

\bibitem{Sun2016CORAL}
Baochen Sun, Jiashi Feng, and Kate Saenko.
\newblock Return of frustratingly easy domain adaptation.
\newblock In {\em Proceedings of the Thirtieth AAAI Conference on Artificial Intelligence}, AAAI'16, page 2058–2065. AAAI Press, 2016.

\end{thebibliography}

\end{document}